\documentclass[10pt,twocolumn,letterpaper]{article}

\usepackage[pagenumbers]{cvpr} 

\usepackage[dvipsnames]{xcolor}

\usepackage[accsupp]{axessibility}
\usepackage{makecell}
\usepackage{multirow}
\usepackage{amssymb}
\usepackage{pifont}
\newcommand{\cmark}{\ding{51}}%
\newcommand{\xmark}{\ding{55}}%

\newcommand*\colourchecksnow[1]{%
  \expandafter\newcommand\csname #1snow\endcsname{\textcolor{#1}{\ding{100}}}%
}
\colourchecksnow{cyan}

\newcommand{\model}{Video ReCap}
\newcommand{\umodel}{Video ReCap-U}
\newcommand{\dataset}{Ego4D-HCap}
\newcommand{\lavila}{LaViLa}

\definecolor{cvprblue}{rgb}{0.21,0.49,0.74}
\usepackage[pagebackref,breaklinks,colorlinks,citecolor=cvprblue]{hyperref}

\def\paperID{523} 
\def\confName{CVPR}
\def\confYear{2024}

\title{Video ReCap: Recursive Captioning of Hour-Long Videos}

\author{Md Mohaiminul Islam$^1$
\quad
Ngan Ho$^1$
\quad
Xitong Yang$^2$
\quad
Tushar Nagarajan$^2$
\\
Lorenzo Torresani$^2$
\quad
Gedas Bertasius$^1$\\
$^1$UNC Chapel Hill \quad\quad $^2$Meta AI
\\
\href{https://sites.google.com/view/vidrecap}{https://sites.google.com/view/vidrecap}
}

\begin{document}
\maketitle
\begin{abstract}
Most video captioning models are designed to process short video clips of few seconds and output text describing low-level visual concepts (e.g., objects, scenes, atomic actions). However, most real-world videos last for minutes or hours and have a complex hierarchical structure spanning different temporal granularities. We propose Video ReCap, a recursive video captioning model that can process video inputs of dramatically different lengths (from 1 second to 2 hours) and output video captions at multiple hierarchy levels. The recursive video-language architecture exploits the synergy between different video hierarchies and can process hour-long videos efficiently. We utilize a curriculum learning training scheme to learn the hierarchical structure of videos, starting from clip-level captions describing atomic actions, then focusing on segment-level descriptions, and concluding with generating summaries for hour-long videos. Furthermore, we introduce Ego4D-HCap dataset by augmenting Ego4D with 8,267 manually collected long-range video summaries. Our recursive model can flexibly generate captions at different hierarchy levels while also being useful for other complex video understanding tasks, such as VideoQA on EgoSchema. Data, code, and models are publicly available at \url{https://sites.google.com/view/vidrecap}.
\vspace{-0.5cm}
\end{abstract}  
\section{Introduction}
\label{sec:intro}

\begin{figure*}
    \vspace{-5mm}
    \centering
    \resizebox{1\textwidth}{!}{%
    \includegraphics[width=1\linewidth]{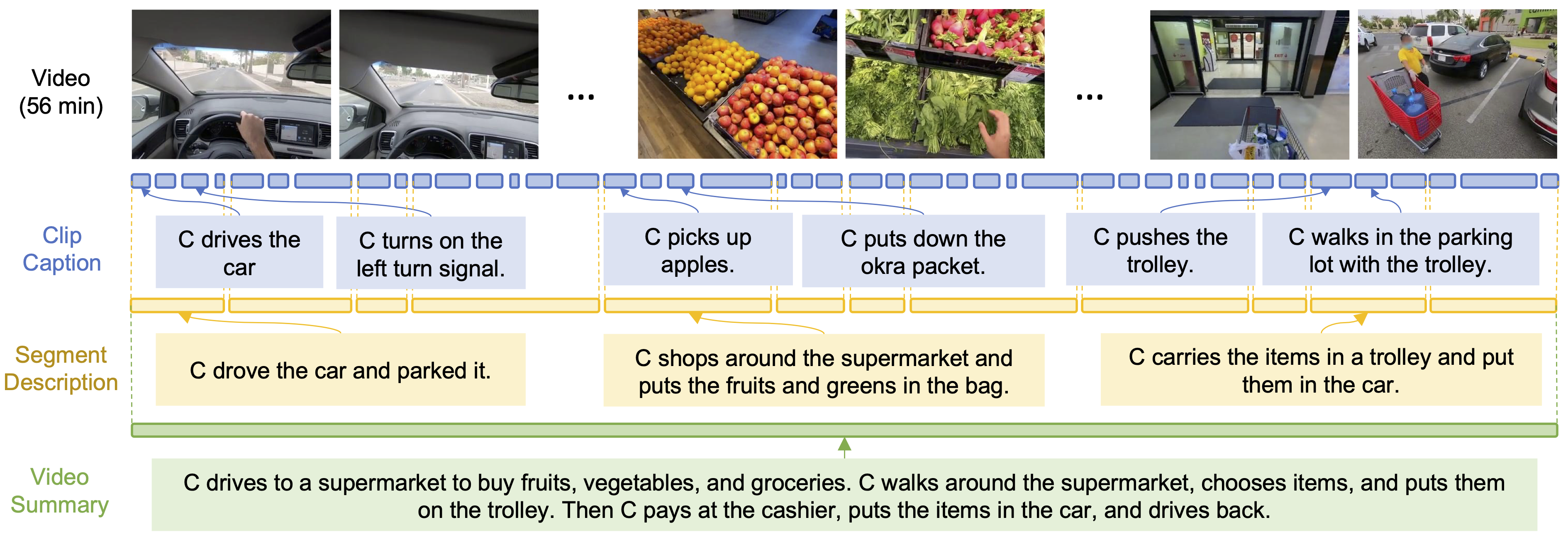}
    }
    \vspace{-0.6cm}
    \caption{\textbf{Hierarchical Video Captioning.} We aim to generate hierarchical captions for a long-range video (e.g., several hours long) at three temporal granularities. First, we generate short clip captions for each few seconds of the video focusing on atomic human actions. Afterward, we produce medium-length segment descriptions for every few minutes of the video, capturing the intermediate steps within a longer activity or a video segment within an extended storyline. Finally, our method generates a summary for a long-range video depicting the overall intent and goals of the actors in the video.\vspace{-0.3cm}}
    \label{fig:teaser} 
\end{figure*}

Many videos in the real world exhibit a hierarchical information structure that spans human behaviors at different temporal granularities (i.e., atomic actions, intermediate activity steps, long-term goals, etc.). However, most modern video captioning models ignore hierarchical video structure and are specifically tailored for short video inputs, typically limited to 5-15 seconds \cite{bahdanau2014neural, rohrbach2013translating, sutskever2014sequence, pan2017video, seo2022end, pei2019memory, hori2017attention, yang2023vid2seq, luo2020univl, sun2019videobert, wang2022omnivl, chen2023valor}. These short-range captioning methods capture atomic actions and low-level visual details, such as objects and scenes. Moreover, these models are often prohibitively resource-intensive when applied to longer videos, making them ill-suited for understanding human activities occurring over long periods (e.g., several hours) \cite{lei2020mart, yang2023vid2seq,sutskever2014sequence, seo2022end}.

In this paper, we investigate a hierarchical video captioning task requiring generating captions at multiple hierarchy levels given a long video input (e.g., several minutes to several hours). Studies in psychology~\cite{barker1954midwest, botvinick2004doing, cooper2006hierarchical} and social cognitive theories~\cite{bandura1999social} have shown the inherent hierarchical structures of human behavior, consisting of atomic actions at the lowest level, intermediate steps in the middle and overall goals/intents at the highest level of the hierarchy. Inspired by these prior studies, we also assume three levels of hierarchies for our video captioning task. At the most granular level, video captions describe individual frames or short video clips of several seconds, focusing on low-level visual elements such as objects, scenes, and atomic actions. As we move up the hierarchy, the short-term captions coalesce into medium-length video segment descriptions spanning activities extending beyond brief moments, such as the intermediate steps within broader activities (e.g., a single step in a cooking recipe) or short segments or sequences within a more extended storyline (e.g., a several minute-long scene within a movie). Lastly, the top level of the hierarchy encapsulates the long-term human goals in the video, intricate relationships between events and characters, and the overarching purpose behind the video, which can be captured via long-range video summaries (See Figure~\ref{fig:teaser}).

The task of hierarchical video captioning poses several technical challenges. Firstly, it necessitates models capable of handling vastly different input lengths, ranging from a few seconds to several hours. This contrasts with most existing methods, designed for fixed video durations of up to a few minutes. Secondly, long-range videos are highly redundant, requiring the model to aggregate only essential information while discarding unimportant visual cues. Thirdly, another critical challenge is comprehending the hierarchical structure in long videos and leveraging the synergy between distinct hierarchies.

To address these technical challenges, we propose \model, a model capable of processing videos of dramatically different lengths where input time spans may differ by up to three orders of magnitude (from a handful of seconds to a few hours) and generating captions at multiple hierarchy levels. Our model encompasses three key attributes that empower its hierarchical video captioning capability. Firstly, \model\ adopts a recursive video-language architecture, allowing it to generate captions across distinct hierarchical tiers. At the first level, the model generates captions from features extracted from short video clips, typically lasting a few seconds. As we move up the hierarchy, the model uses sparsely sampled video features and captions generated at the previous hierarchy level as inputs to produce video captions for the current hierarchy level. Such a recursive design effectively leverages the synergy between different video hierarchies and allows us to handle very long video inputs (e.g., up to 2 hours) efficiently. Moreover, it facilitates our model to leverage the powerful reasoning abilities of modern LLMs. Secondly, we implement a curriculum learning scheme, commencing with training on short video clip captions and progressively incorporating data from higher-level hierarchies, namely medium-length segment descriptions and long-range video summaries. Such a hierarchical curriculum learning strategy allows the model to gradually learn the hierarchical structure of the video, starting from short low-level captions to long high-level video summaries. Thirdly, to mitigate the challenge of limited manually annotated hierarchical captioning data, we use LLMs to generate pseudo-summary data spanning different temporal lengths and then use these pseudo-annotations as additional data to train our model.

To evaluate \model, we introduce \dataset\ dataset, a new hierarchical video captioning benchmark that contains long-range egocentric videos lasting up to several hours with manually annotated captions at multiple hierarchical levels. To build \dataset\ benchmark, we utilize Ego4D~\cite{grauman2022ego4d}, the largest publicly available long-range egocentric video dataset, which provides time-stamped captions and video-segment summaries of up to 5 minutes. We then augment the subset of Ego4D videos with manually annotated 8,267 long-range video summaries, where each video spans up to two hours. Consequently, the \dataset\ becomes a rich resource with three levels of hierarchical captions for long untrimmed egocentric videos, encompassing captions for short clips, intermediate descriptions for few-minute video segments, and video-level summaries for long video sequences.

Our results show that \model~outperforms strong prior video captioning baselines~\cite{zhao2023learning, li2023blip} across all three temporal hierarchies by a large margin. We also demonstrate that \model~can be effectively used for other complex video understanding tasks, such as long-form video question-answering on EgoSchema~\cite{mangalam2023egoschema} where our approach outperforms the previous best method by a substantial margin (\textbf{+18.13\%}). 

\section{Related Works}
\label{sec:related}

\vspace{1mm}
\noindent\textbf{Video Captioning Methods.} Early works in video captioning used template-based approaches~\cite{lan2017fluency, kojima2002natural, rohrbach2013translating, sun2014semantic, xu2015jointly}. Subsequently, these methods were replaced by deep learning methods built using CNN-RNN encoder-decoder architectures~\cite{yao2015describing, donahue2015long, venugopalan2015sequence, pan2016hierarchical, pan2017video, baraldi2017hierarchical, song2018deterministic, wang2018reconstruction}. The recent introduction of Transformer~\cite{vaswani2017attention, dosovitskiy2020image} led to a plethora of transformer-based video captioning methods~\cite{pei2019memory, hori2017attention, pan2017video, baraldi2017hierarchical, song2018deterministic, wang2018reconstruction, lei2020mart, yang2023vid2seq,sutskever2014sequence, seo2022end}. Though these approaches have shown great success in short clip captioning, most are limited to short videos of a few seconds and cannot generate captions spanning multiple temporal hierarchies for hour-long videos. 

\vspace{1mm}
\noindent\textbf{Video Captioning Datasets.} Most existing video captioning datasets contain short video clip inputs (5-30 seconds)~\cite{chen2011collecting, xu2016msr, wang2019vatex, rohrbach2017movie}. There exist several datasets with longer videos of 1-5 minutes~\cite{zhou2018towards,krishna2017dense, huang2020multimodal}, but the captions of these datasets still focus on short-term visual concepts (e.g., atomic actions, presence of objects, etc.). Instead, our work aims to develop models and datasets for hierarchical video captioning that spans multiple temporal granularity levels ranging from short clip captions to long-range video summaries. To do this, we introduce \dataset~dataset by augmenting Ego4D with long-range video summaries of hour-long videos. This leads to a hierarchical video captioning dataset consisting of short clip captions, medium-range segment descriptions, and long-range video summaries.

\noindent\textbf{Hierarchical Video Understanding.} Several recent datasets include hierarchical activity annotations for procedural videos~\cite{tang2019coin, zhukov2019cross, sener2022assembly101, bansal2022my, song2023ego4d}. However, these datasets define a fixed taxonomy for the activity labels of each hierarchy and focus on procedural activity recognition. In contrast, we assume free-form natural language descriptions for multiple levels to capture inherent hierarchical structure in real-world videos (not limited to only instructional videos). Aside from the datasets, several methods~\cite{ashutosh2023hiervl, Zhang2018CrossModalAH, Li2020HeroHE} learn hierarchical feature embeddings for several-minute-long videos (e.g., 5 minutes). In contrast, our work focuses on generating free-form hierarchical captions for hour-long videos at multiple temporal scales.

\section{Technical Approach}
\label{sec:techincal}

\begin{figure*}
    \vspace{-5mm}
    \centering
    \includegraphics[width=1\linewidth]{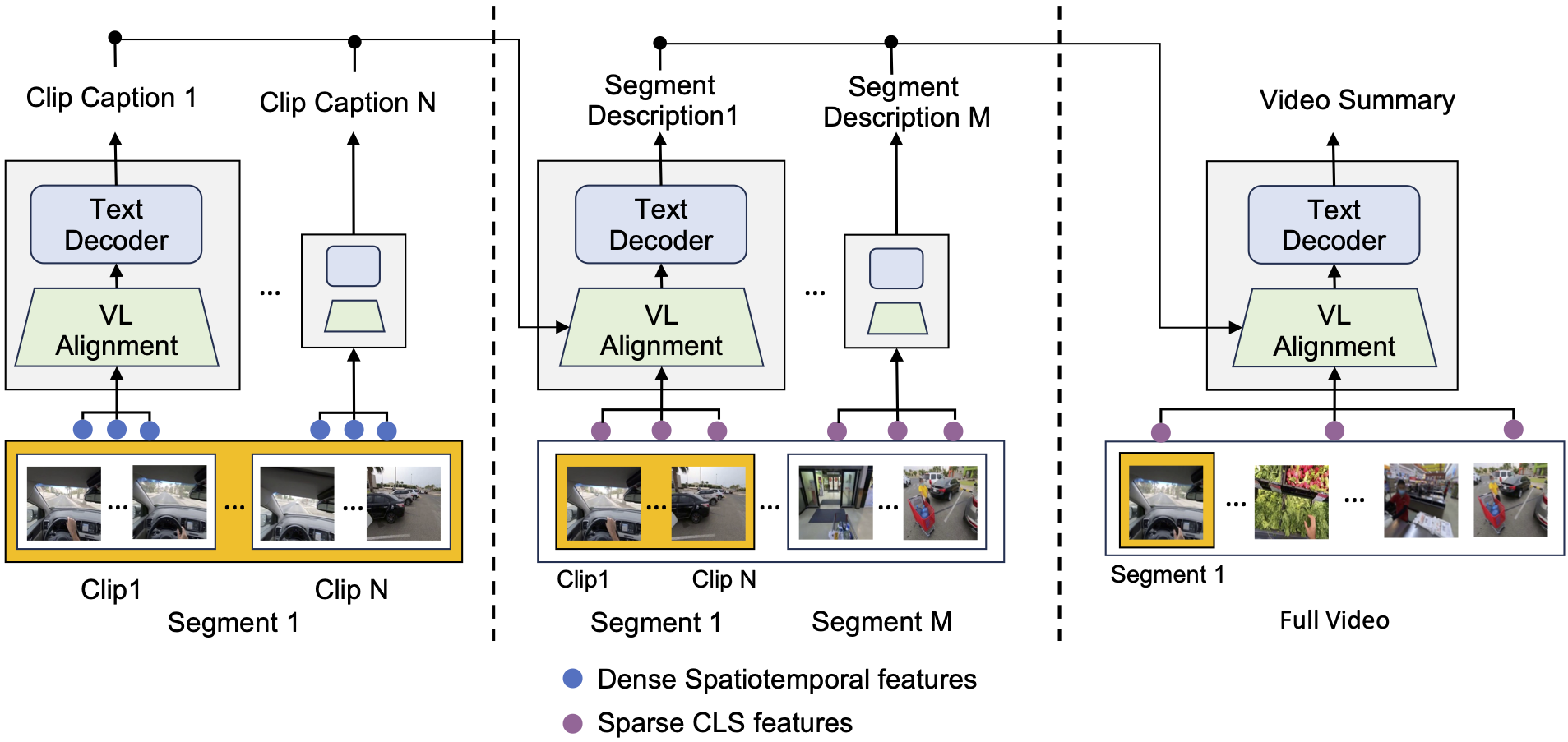}
    \vspace{-0.6cm}
    \caption{\textbf{The \model\ model}. \textbf{(Left)} First, we generate captions for each short clip (e.g., a few seconds long) of the video using the dense spatiotemporal features extracted by a pretrained video encoder (not shown in the figure). \textbf{(Middle)} Then \model\ produces segment descriptions for every few minutes of the video using sparsely sampled features (e.g., CLS features) and the previously generated clip captions belonging to a particular segment. \textbf{(Right)} Finally, \model\  generates the full video summary by utilizing sparsely sampled CLS features from the entire video and the previously generated segment descriptions. The Video-Language (VL) Alignment module maps the video and text features to a joint space so that the subsequent text decoder can jointly process them. Note: the yellow box represents the first segment of the video in each of the three panels, zooming in from right to left.\vspace{-0.3cm}}
    \label{fig:model} 
\end{figure*}

\subsection{Problem Overview}
\label{sec: task definition}

Given a long, untrimmed video input, we aim to generate textual captions at multiple hierarchy levels of the video. Formally, as our inputs, we consider a long-range video sequence $V_i=[I_i^{(t)}]_{t=1,\hdots, T}$ comprised of $T$ RGB frames, denoted by $I_i^{(t)}$. Our goal is then to generate captions at three distinct hierarchical levels: $Y_i^{(\ell)} = [y_{i,j}^{(\ell)}]_{j=1,\hdots,|Y_i^{(\ell)}|}$ for $\ell=1,2,3$, where $y_{i,j}^{(\ell)}$ depicts a $j^{th}$ word in a caption $i$ for the hierarchy level $l$. Each hierarchy of captions is generated sequentially starting with the short-term video clip captions, $Y_i^{(1)}$,  describing fine-grained actions and objects occurring within few seconds intervals throughout the video (e.g., a person picks up an apple in Figure~\ref{fig:teaser}). Afterward, the model outputs medium-length segment descriptions $Y_i^{(2)}$, which capture intermediate steps or summaries unfolding over a few minutes of the video (e.g., a person driving a car and parking it in Figure~\ref{fig:teaser}). Finally, the model finishes its generation with long-range video summaries $Y_i^{(3)}$ representing video content for the entire video input. 

\subsection{Recursive Video-Language Model}
\label{sec: model architecture}

We now describe the \model\ model, which contains three high-level components: a Video Encoder, Video-Language Alignment, and a Recursive Text Decoder. We illustrate our approach in Figure~\ref{fig:model} and describe each component below.

\vspace{1mm}
\noindent\textbf{Video Encoder.} First, we utilize an off-the-shelf video encoder (e.g., TimeSformer~\cite{bertasius2021space}) to extract features from a long-range video. Given a short video clip, the video encoder outputs dense spacetime features. We divide the entire video uniformly and extract a sequence of features $X_i=[x_{i,j}]_{j=1,\hdots,|C|}$, where $|C|$ is the number of video clips, $x\in \mathbb{R}^{F \times H \times W \times D}$ is the spatiotemporal features of a particular clip, $F$ is the number of frames, $H$ is the height, $W$ is the width, and $D$ is the feature dimension. We use dense spacetime features for short-clip captions so that the model can identify low-level visual cues (i.e., objects and atomic actions); for higher-level captions (e.g., segment descriptions and video summaries), we use global features (e.g., CLS features) to reduce the computational cost and capture the global properties of long video inputs.

\vspace{1mm}
\noindent\textbf{Video-Language Alignment.} Next, we utilize a Video-Language (VL) Alignment module which takes the video features, $X_i$ and the captions generated in the previous hierarchy $Y_i^{(\ell-1)}$ as input and outputs a fixed number of embeddings $Z_i=[z_{i,j}]_{j=1,\hdots,|Z|}$, where $z\in \mathbb{R}^{D_z}$, $|Z|$ is the number of embeddings, and $D_z$ is the hidden dimension. The objective of the alignment module is to map the video and text features to the joint feature space so that the subsequent text decoder can jointly process both features as in~\cite{li2023blip}. Moreover, this scheme enables us to compress a large number of video and text features (e.g., several thousand) into a small set of embeddings (e.g., 256), dramatically reducing the computational cost. In particular, we use a frozen pre-trained language model (e.g., DistilBERT~\cite{sanh2019distilbert}) to learn a fixed number of video embeddings from the video features $X_i$ by injecting trainable cross-attention layer inside each transformer block of the LM. We also learn a fixed number of text embeddings from the captions generated at the previous hierarchy $Y_i^{(\ell-1)}$ by using a similar frozen LM with trainable cross-attention layers. Finally, we concatenate the video and text embeddings to get the joint embeddings $Z_i$, which is used by the subsequent text decoder for generating captions $Y_i^{(\ell)}$. Note that the first hierarchy level (i.e., clip caption) has no text features and uses only video embeddings as $Z_i$.

\vspace{1mm}
\noindent\textbf{Recursive Text Decoder.} We use a pretrained language model (e.g., GPT2~\cite{radford2019language}) as our recursive text decoder for generating captions at multiple hierarchy levels. The decoder takes the video-text embeddings $Z_i$ produced by the video-language alignment module (described above) and then generates captions $Y_i^{\ell}$ for the hierarchy $\ell$. Note that we use captions generated at the previous hierarchy level $Y_i^{\ell -1}$ as one of the inputs (along with video features $X_i$), which enables a recursive caption generation pipeline. Note that for short-term caption generation (i.e., $Y_i^{1}$), the textual feature set is initialized as empty (i.e., the base case of our model's recursion). Following prior works~\cite{zhao2023learning, alayrac2022flamingo}, we insert trainable cross-attention blocks inside each transformer layer of our textual decoder and freeze the remaining layers. The cross-attention layer attends to video-text embeddings of the alignment module. Therefore, the proposed \model\ models the likelihood of caption $Y^{(\ell)}$ conditioned on the video $X$ and the captions generated at lower-level hierarchy $Y^{(\ell -1)}$ using the following training objective:

\vspace{-0.2cm}
\begin{align}
p(Y^{(\ell)} | X) = \prod_{k=1}^K p(y_k^{(\ell)} | y^{(\ell)}_{< k}, X, Y^{(\ell -1)})
\label{eq:model}
\end{align}

Here, $y_k^{(\ell)}$ denotes the language token of the caption, $y^{(\ell)}_{< k}$ is the set of preceding tokens, and $Y^{(0)} = \emptyset$.

\subsection{Hierarchical Curriculum Learning}
\label{sec: curriculum}

\begin{figure}
    \centering
    \resizebox{0.5\textwidth}{!}{%
    \includegraphics[width=1\linewidth]{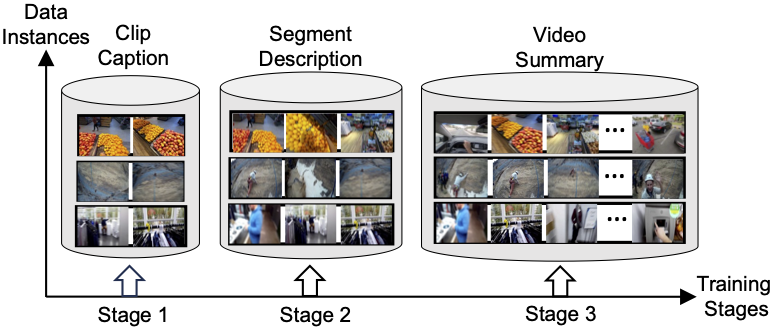}
    }
    \vspace{-0.6cm}
    \caption{\textbf{Hierarchical Curriculum Learning.} We gradually learn the hierarchical structure of the video, starting from short low-level captions to long high-level video summaries.\vspace{-0.6cm}}
    \label{fig:curriculum} 
\end{figure}

Training a recursive video-language model is challenging for several reasons. First, the model must process videos of dramatically different input lengths (i.e., from a few seconds to several hours). Second, there is a significant data imbalance where short-term clip captions vastly outnumber the number of video segment descriptions and long-range summaries. Finally, exploiting the synergy between different hierarchy levels is crucial for generating meaningful and contextually relevant captions. To overcome these challenges, we draw motivation from classic studies of psychology~\cite{barker1954midwest, botvinick2004doing, cooper2006hierarchical, bandura1999social}, which show a hierarchical organization of human perception of actions. Just as humans first perceive atomic actions before grasping mid-level actions and then infer goals from mid-level activities, our training strategy unfolds in a similar hierarchical fashion. Specifically, our training begins with samples from the lowest hierarchy level, namely clip captions. Subsequently, we train our model with higher-level captions, e.g., medium-length segment descriptions and long-range video summaries. This strategic progression allows the model to gradually understand the intricate hierarchical structure inherent in videos and maximize the synergy between all hierarchies. Moreover, this strategy effectively handles highly imbalanced training data across different hierarchies. Figure~\ref{fig:curriculum} shows an overview of the proposed curriculum learning strategy.

\subsection{Additional Supervision using Language Models}
\label{sec: llm}

\begin{figure}
    \centering
    \resizebox{0.5\textwidth}{!}{%
    \includegraphics[width=1\linewidth]{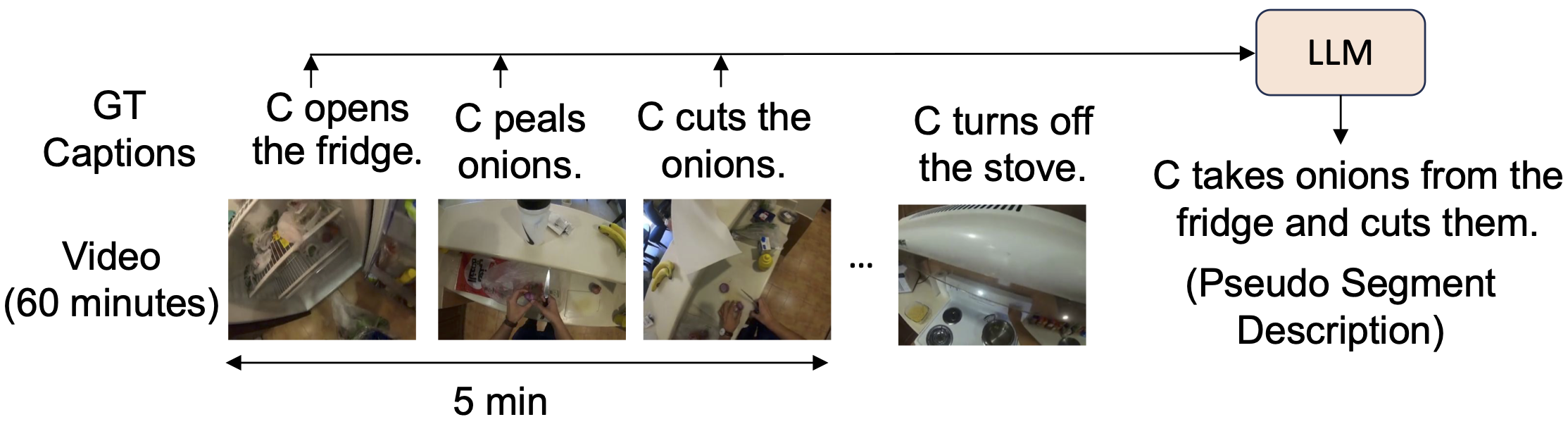}
    }
    \vspace{-0.7cm}
    \caption{\textbf{Large Language Model Supervision.} Given short-term ground truth captions, we use an LLM to generate pseudo-ground truth annotations for medium-length segment descriptions and long-range video summaries to augment our training data.\vspace{-0.4cm}} 
    \label{fig:llm_sup} 
\end{figure}

Collecting captioning annotations for hour-long videos is time-consuming and costly. Thus, another critical challenge associated with hierarchical video captioning is the scarcity of manually annotated hierarchical captioning data, particularly for medium-length segment descriptions and long-range video summaries. We leverage Large Language Models (LLMs) to mitigate this issue. LLMs can effectively incorporate information from text inputs of varying lengths, which aligns perfectly with our objective of guiding the video model to generate captions across multiple hierarchies. Motivated by these insights, we use LLMs to generate a large number of pseudo-caption annotations for medium-length and long-range videos (i.e., our last two hierarchies). The process involves two main steps. First, given manually annotated hierarchical captions, we finetune an LLM teacher to generate medium-length segment descriptions and long-range video summaries from short-term clip captions concatenated across varying temporal durations. Afterward, we use such LLM-generated pseudo ground truth caption data as additional training samples to train \model\ (see Figure~\ref{fig:llm_sup}). Our experiments indicate that such pseudo ground truth data generated by LLMs effectively complements manually annotated data and significantly improves our model's captioning ability.  

\subsection{Implementation Details}

We use TimeSformer~\cite{bertasius2021space} as our video encoder to extract features that take an input clip of $4$ RGB frames of $224\times 224$. We use GPT2~\cite{radford2019language} as our default text-decoder, with a hidden dimension of $768$ and $12$ transformer blocks. We use Adam optimizer~\cite{Kingma2014AdamAM} with a learning rate of $3^{-5}$ and a weight decay of $0.01$. Our training pipeline also utilized cosine scheduling strategy~\cite{Loshchilov2017DecoupledWD}. Please refer to supplementary materials for additional implementation details.
\section{\dataset~Dataset}

\begin{table}[t]
\centering
\begin{tabular}{ccc}
\toprule
Hierarchy Level     & \# Samples & Avg. Duration \\
\toprule
Clip Caption        & 5.27M      & 0.96 sec    \\
Segment Description & 17.5K     & 2.87 min     \\
Video Summary       & 8.3K      & 28.46 min   \\
\bottomrule
\end{tabular}
\vspace{-0.3cm}
\caption{\textbf{Summary of \dataset\ dataset.}\vspace{-0.4cm}}
\label{tab:datset}
\end{table}

We now describe our introduced \dataset\ dataset, a hierarchical video captioning dataset comprised of a three-tier hierarchy of captions: short clip-level captions, medium-length video segment descriptions, and long-range video-level summaries. To construct \dataset, we leverage Ego4D~\cite{grauman2022ego4d}, the largest publicly available egocentric video dataset. Ego4D videos have several unique features, making them ideal for the hierarchical video captioning task. First, most videos in Ego4D are orders of magnitude longer (e.g., several hours) than the traditional video captioning datasets. Second, egocentric videos typically contain goal-driven and human activities at different hierarchy levels. Third, Ego4D videos capture human behaviors from various scenarios such as cooking, gardening, assembly, etc.

While Ego4D comes with time-stamped atomic captions and video-segment descriptions spanning up to 5 minutes, it lacks video-level summaries for longer video durations. To address this issue, we annotate a subset of 8,267 Ego4D videos with long-range video summaries, each spanning up to two hours. This enhancement provides a three-level hierarchy of captions, making it a perfect resource for validating the effectiveness of our model on the hierarchical video captioning task. In Table~\ref{tab:datset}, we provide a detailed summary of our introduced \dataset\ subset. Please refer to our supplementary materials for a more detailed analysis of the \dataset\ dataset. 

\begin{figure*}
    \centering
    \resizebox{1\textwidth}{!}{%
    \includegraphics[width=1\linewidth]{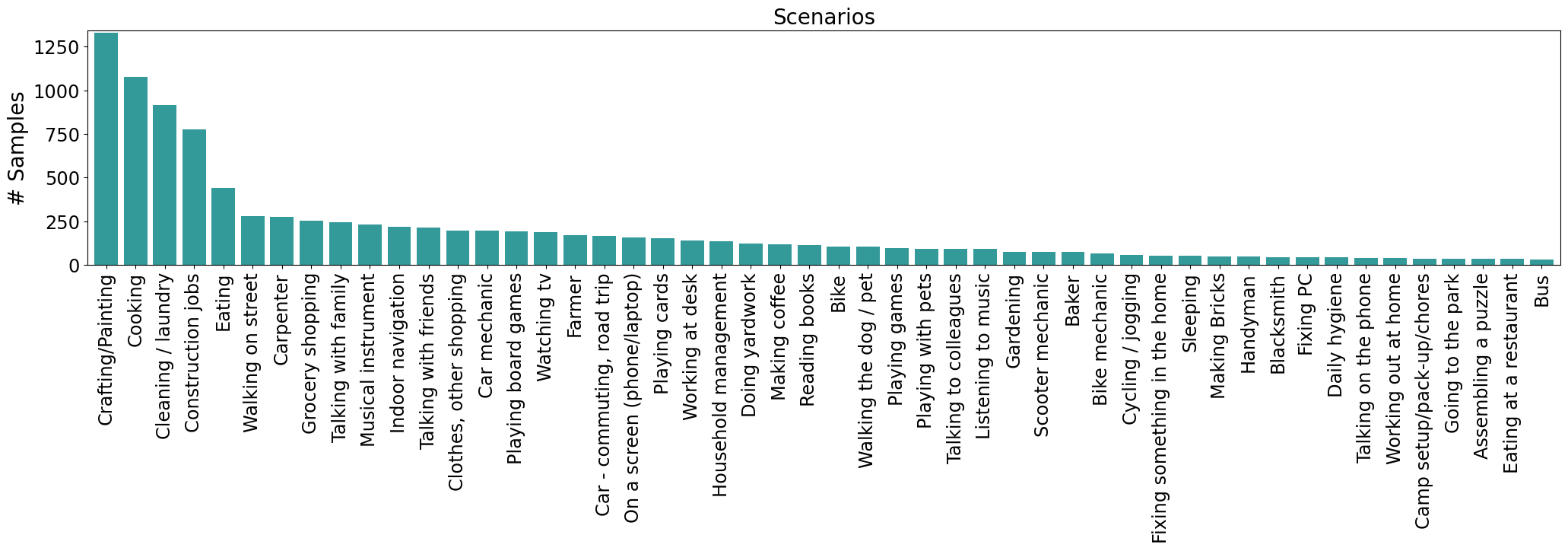}
    }
    \caption{\textbf{Distribution of the most common 50 scenarios in \dataset\ dataset.}}
    \label{fig:data scenarios} 
\end{figure*}

\begin{figure*}
    \centering
    \resizebox{1\textwidth}{!}{%
    \includegraphics[width=1\linewidth]{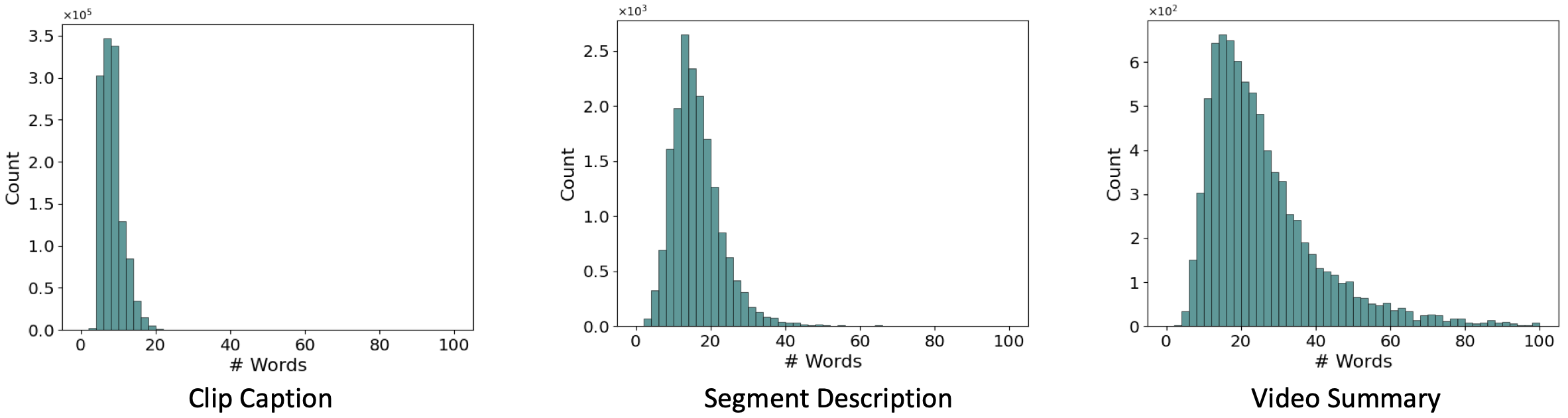}
    }
    \caption{\textbf{Distribution of the lengths of three hierarchical captions of the \dataset\ dataset.}}
    \label{fig:data lengths} 
\end{figure*}

Our proposed \dataset\ dataset contains videos that capture diverse scenarios in various contexts, such as household settings, outdoor environments, workplaces, leisure activities, and more, totaling 127 distinct scenarios. The distribution of the most common 50 scenarios is illustrated in \Cref{fig:data scenarios}. The distribution of caption lengths for three hierarchy levels in the \dataset\ dataset is illustrated in \Cref{fig:data lengths}. Notably, clip captions are generally shorter, averaging 7.74 words per caption. In comparison, segment descriptions display a medium length, averaging 15.79 words, while video summaries are the longest, with an average of 25.59 words. Additionally, we observe that the maximum length for a clip caption is 43 words, while segment descriptions and video summaries can extend to 73 and 172 words, respectively. Our supplementary materials include more details on the dataset and our annotation collection process.

\section{Experimental Setup}
\label{sec:exp setup}

\begin{table*}[t]
\centering
    \vspace{-5mm}
    \begin{subtable}{0.7\linewidth}
    \centering
    \resizebox{1\textwidth}{!}{%
    \begin{tabular}{c|ccc|ccc}
    \toprule
    \multirow{2}{*}{Model} & \multirow{2}{*}{\begin{tabular}[c]{@{}c@{}}Visual\\ Encoder\end{tabular}} & \multirow{2}{*}{\begin{tabular}[c]{@{}c@{}}Text \\ Decoder\end{tabular}} & \multirow{2}{*}{\begin{tabular}[c]{@{}c@{}}Train\\ Params\end{tabular}} & \multicolumn{3}{c}{Clip Caption}                 \\ \cline{5-7}  &                                                                           &                                                                          &                                                                         & CIDEr          & ROUGE-L        & METEOR         \\ \hline
    \textbf{Zero-Shot}\\
    BLIP2~\cite{li2023blip}                  & VIT-G                                                                     & FT5-XL                                                                   & 0                                                                       & 8.1            & 7.4            & 12.7           \\
    \textbf{Finetuned}\\
    \lavila~\cite{zhao2023learning}                 & TSF-B                                                                     & GPT2                                                                     & 258M                                                                     & 88.56          & 47.64          & 28.03          \\
    \model             & TSF-B                                                                     & GPT2                                                                     & 339M                                                                    & \textbf{98.35} & \textbf{48.77} & \textbf{28.28} \\
    \umodel      & TSF-B                                                                     & GPT2                                                                     & 113M                                                                    & 92.67          & 47.90          & 28.08          \\ \hline
    \end{tabular} 
    }
    \caption{\small{Results for short-range clip captioning.}}
    \vspace{1.5mm}
    \end{subtable}
    \begin{subtable}{1\linewidth}
    \centering
    \resizebox{1\textwidth}{!}{%
    \begin{tabular}{c|cccc|ccc|ccc}
    \hline
    \multirow{2}{*}{Model} & \multirow{2}{*}{\begin{tabular}[c]{@{}c@{}}Video\\ Encoder\end{tabular}} & \multirow{2}{*}{\begin{tabular}[c]{@{}c@{}}Text \\ Decoder\end{tabular}} & \multirow{2}{*}{\begin{tabular}[c]{@{}c@{}}Train\\ Params\end{tabular}} & \multirow{2}{*}{\begin{tabular}[c]{@{}c@{}}Pseudo\\ Ann.\end{tabular}} & \multicolumn{3}{c|}{Segment Description}         & \multicolumn{3}{c}{Video Summary}                \\ \cline{6-11} 
                           &   &                                                                          &                                                                         &                                                                        & C              & R              & M              & C              & R              & M              \\ \hline
    \textbf{Zero-Shot}\\
    BLIP2~\cite{li2023blip} + GPT3.5~\cite{brown2020language}         & VIT-G                                                                    & FT5-XL                                                                   & 0                                                                       &    \xmark                                                                    & 5.68           & 16.87          & 13.47          & 11.13          & 22.41          & 12.10          \\
    \lavila~\cite{zhao2023learning} + GPT3.5~\cite{brown2020language}        & TSF-B                                                                    & GPT2                                                                     & 0                            &     \xmark                                                                                                           & 5.79           & 19.77          & 13.45          & 12.16          & 24.49          & 12.48          \\
    \textbf{Finetuned}\\
    \lavila~\cite{zhao2023learning} + GPT2~\cite{radford2019language}          & TSF-B                                                                    & GPT2                                                                     & 336M                                                                    &  \xmark                                                                      & 38.22          & 38.10          & 16.58          & 17.98          & 29.48          & 12.81          \\
    \lavila~\cite{zhao2023learning} + FLANT5~\cite{chung2022scaling}        & TSF-B                                                                    & FT5-XL                                                                   & 586M                                          &                \xmark                                                                               & 39.13          & 38.77          & 16.88          & 20.12          & 30.06          & 13.17          \\
    \lavila~\cite{zhao2023learning}                 & TSF-B                                                                    & GPT2                                                                     & 258M                                            &     \xmark                                                                                  & 24.63          & 33.31          & 15.30          & 6.54           & 23.97          & 10.95          \\
    \model             & TSF-B                                                                    & GPT2                                                                     & 339M                                &                     \xmark                                                                                & 41.74          & 39.04          & 18.21          & 28.06          & 32.27          & 14.26          \\
    \model             & TSF-B                                                                    & GPT2                                                                     & 339M                                                                    &       \cmark                                                                 & \textbf{46.88} & \textbf{39.73} & \textbf{18.55} & 29.34          & 32.64          & \textbf{14.45} \\
    \umodel      & TSF-B                                                                    & GPT2                                                                     & 113M                                    &            \cmark                                                                                       & 45.60          & 39.33          & 18.17          & \textbf{31.06} & \textbf{33.32} & 14.16  \\ \bottomrule
    \end{tabular}
    }
    \caption{\small{Results for medium-length segment description and long-range video summary generation.}}
    \end{subtable}
    \vspace{-6mm}
    \caption{\textbf{Main Results on the \dataset\ dataset.} All results are evaluated in standard CIDEr (C), ROUGE-L (R) and METEOR (M) metrics. We observe several interesting trends. First, finetuned methods perform significantly better than the zero-shot baselines. Second, the \model\ model achieves the best results in video captioning across all three hierarchies, surpassing strong prior baselines such as \lavila~\cite{zhao2023learning}. Third, using LLM-generated pseudo annotations leads to a significant boost in performance. Lastly, the unified variant of the model produces competitive results while having a significantly smaller number of trainable parameters than our standard variant.\vspace{-4mm}}
    \label{tab:main results}
\end{table*}

\subsection{Hierarchical Video Captioning Baselines}
Hierarchical video captioning is a relatively unexplored task, so there are no well-established baselines for comparing our work. Thus, we introduce the following video-language baselines, which we extend for this task.

\begin{itemize}
    \item \textbf{Zero-Shot Baselines:}
    \begin{enumerate}
        \item \textbf{BLIP2}~\cite{li2023blip}. A zero-shot baseline for \textit{short-term clip captioning} that utilizes a state-of-the-art image captioning model.
        
        \item \textbf{BLIP2 + GPT3.5}~\cite{li2023blip, brown2020language}. A zero-shot text-based baseline for \textit{video segment descriptions} and \textit{long-range video summaries}. Given BLIP2-generated captions, it uses GPT3.5 to generate video segment descriptions and long-range video summaries.
        
        \item \textbf{\lavila\ + GPT3.5}~\cite{zhao2023learning, brown2020language}. Similar to the above, a zero-shot baseline for \textit{video segment} and \textit{summary} generation using LaViLa captions fed into GPT3.5.
    \end{enumerate}

    \vspace{1mm}
    \item \textbf{Finetuned Baselines:}
    \begin{enumerate}
        \item \textbf{\lavila\ + GPT2}~\cite{zhao2023learning, radford2019language}. A fully-finetuned text-based baseline that takes LaViLa-generated clip captions and finetunes a text-only GPT2 model for \textit{segment description} and \textit{video summary} generation while keeping the underlying LaViLa model frozen.
        
        \item \textbf{\lavila\ + FLAN-T5}~\cite{zhao2023learning, chung2022scaling}. Similar to the above, a fully-finetuned text-based baseline that uses FLAN-T5 rather than GPT2 for \textit{segment description} and \textit{video summary} generation.

        \item \textbf{\lavila}~\cite{zhao2023learning}. A video-based baseline, finetuned end-to-end to generate \textit{short-term captions, medium-length segment descriptions}, and \textit{long-range video summaries} directly using video inputs. Note that this baseline uses the same video encoder, text decoder, and other experimental settings as our model.  
    \end{enumerate}

\end{itemize}

\subsection{Our Model Varients}

\begin{enumerate}
    \item \textbf{\model .} This variant of our model uses a shared video encoder but separate text decoders and video-language alignment modules to generate captions at different hierarchy levels (i.e., the weights across different hierarchies are not shared). Due to the increased model capacity of having specialized modules for each hierarchy, this variant typically produces the best performance. 
    \item \textbf{\model -U.} The unified variant using shared parameters across all hierarchies. Since it has a lot fewer trainable parameters than the previous variant, it is more efficient but performs slightly worse in certain settings.
\end{enumerate}

\section{Results and Analysis}
\label{sec:results}

\subsection{Hierarchical Video Captioning Results}
\label{sec:main results}

In Table~\ref{tab:main results}, we present our main results for hierarchical video captioning.  We use standard captioning metrics, including CIDEr~\cite{vedantam2015cider}, ROUGE-L~\cite{lin2004rouge}, and METEOR~\cite{banerjee2005meteor} to evaluate our model on the hierarchical video captioning task. Based on these results, we observe several interesting trends. First, we note that zero-shot baselines (e.g., BLIP2~\cite{li2023blip}, BLIP2 + GPT3.5~\cite{brown2020language}, \lavila\ + GPT3.5)  perform considerably worse than the fully finetuned approaches (e.g., \lavila~\cite{zhao2023learning}, \lavila\ + GPT2~\cite{radford2019language}, \lavila\ + FLAN-T5~\cite{chung2022scaling}), underscoring the significance of in-domain learning on the \dataset\ dataset. Second, we observe that the best performing fully-finetuned text-based baseline \lavila\ + FLAN-T5~\cite{chung2022scaling} falls short of our model by 2.61\% CIDEr on video segment description and 9.94\% CIDEr on video summary generation, despite using significantly more trainable parameters (586M vs 339M). This indicates the benefits of using hierarchical video and text inputs rather than just text for video segment description and long-range video summary generation. Third, we notice that our best performing \model\ variant significantly improves upon the strong LaViLa baseline on clip captioning for Ego4D~\cite{grauman2022ego4d}, outperforming it by 9.79\% CIDEr while employing the same visual encoder, text decoder, and training data as our model. We note that while \lavila~uses a transformer resampler~\cite{zhao2023learning, alayrac2022flamingo}, our model utilizes a Language Model-based alignment module (see \Cref{sec: model architecture}), which we found very effective for this particular task.

We also note that the performance of \lavila\ drops significantly for segment description and video summary generation, indicating its inability to handle long-range videos. In contrast,  \model\ maintains strong performance on these longer video inputs, outperforming \lavila\ by 17.11\% CIDEr on segment description and 21.52\% CIDEr on video summary generation. We also note that while \model\ uses more training parameters than \lavila~(258M vs. 339M), \umodel\ has significantly fewer training parameters (113M) than \lavila\ but still outperforms \lavila\ by substantial margins (+20.97\% and +24.50\% in CIDEr for segment description and video summary generation respectively). This indicates that the performance gain of our model comes from the recursive and hierarchical design and not from the larger capacity of the model. Our results also indicate that our model's performance can be further improved (5.14\% CIDEr in segment description and 1.28\% CIDEr in video summary) by incorporating LLM-based supervision (see Section~\ref{sec: llm}). Lastly, the last two rows of Table~\ref{tab:main results} highlight the trade-off between the two variants of our model, i.e., \model\ achieves the highest performance across two out of three hierarchies, while the unified variant, \model -U, attains the second-best performance with significantly fewer trainable parameters.

\subsection{Long-Range VideoQA on EgoSchema}
\label{sec:results es}

\begin{table}[t]
\centering
\Large
\resizebox{0.5\textwidth}{!}{%
\begin{tabular}{cccc}
\toprule
Model               & \begin{tabular}[c]{@{}c@{}}Input \\ Feature\end{tabular} & \begin{tabular}[c]{@{}c@{}}Ego4D\\ Pretrain\end{tabular} & \begin{tabular}[c]{@{}c@{}}QA\\ Acc\end{tabular} \\
\toprule
Random              & -                                     & \xmark                                                         & 20.0                                             \\
GPT3.5~\cite{brown2020language}              & Question                                     & \xmark                                     & 19.57                            \\
\midrule
FrozenBiLM~\cite{Yang2022ZeroShotVQ}          & Video                                                    & \xmark                                                        & 26.9                                             \\
VIOLET~\cite{Fu2022AnES}              & Video                                                    & \xmark                                                         & 19.9                                             \\
mPLUG-Owl~\cite{Ye2023mPLUGOwlME}           & Video                                                    & \xmark                                                         & 31.1                                             \\
InternVideo~\cite{wang2022internvideo}         & Video                                                    & \xmark                                                         & 32.1                                             \\
\midrule
EgoVLP~\cite{lin2022egocentric}             & Video                                                    & \cmark                                                        & 34.86                                            \\
EgoVLPv2~\cite{pramanick2023egovlpv2}            & Video                                                    & \cmark                                                        & 34.12                                            \\
\midrule
LaViLa~\cite{zhao2023learning} + GPT3.5~\cite{brown2020language}     & Captions                                                  & \cmark                                                        & 44.27                             \\
\model~+ GPT3.5~\cite{brown2020language} & Captions                                                  & \cmark                                                        & 46.03                                            \\
\model~+ GPT3.5~\cite{brown2020language} & Hier. Captions                                           & \cmark                                                        & \textbf{50.23}                                            \\
\bottomrule
\end{tabular}
}
\vspace{-0.3cm}
\caption{\textbf{Long-Range VideoQA on EgoSchema~\cite{mangalam2023egoschema}} Our approach achieves state-of-the-art results, outperforming the previous best method, InternVideo, by a substantial margin of 18.13\%. Furthermore, leveraging the hierarchical captions produced by our model leads to 4.2\% and 5.96\% boost in performance compared to short-clip captions generated by \model\ or \lavila~\cite{zhao2023learning}. 
\vspace{-0.4cm}}
\label{tab:egoschema}
\end{table}

In Table~\ref{tab:egoschema}, we validate the effectiveness of our hierarchical video model on the recently introduced long-range video question-answering (VideoQA) EgoSchma dataset~\cite{mangalam2023egoschema}. EgoSchema contains over 5K human-curated multiple-choice question-answer pairs spanning 250 hours of real-world videos, requiring hierarchical reasoning over long videos. We use a simple two-stage approach to perform VideoQA on EgoSchema. First, given long EgoSchema video inputs, we generate hierarchical video captions like before. Afterward, we feed our generated hierarchical video captions as inputs to a text-only GPT3.5~\cite{brown2020language} and prompt it to answer a question about a given video in a zero-shot manner. The simple framework performs very well on this benchmark despite the simplicity. We first observe that compared to the variant of our method that uses only short-term captions as inputs to GPT3.5, the variant that uses hierarchical video captions achieves a significant 4.2\% boost in performance. We also compare our method with a similar baseline that uses LaViLa-generated short-term captions rather than our hierarchical video captions as inputs to GPT3.5 and show that our approach outperforms this baseline by 5.96\%. This highlights the benefits of hierarchical video cues for long-range videoQA. Our results also indicate that our method outperforms the previous best model, InternVideo~\cite{wang2022internvideo} by a large margin of 18.13\%, setting a new state-of-the-art on this benchmark. We note, however, that since InternVideo was never pretrained on Ego4D, the comparison with our approach might be somewhat unfair. Thus, in our comparisons, we also include two recent methods, pretrained on Ego4D, EgoVLP~\cite{lin2022egocentric} and EgoVLPv2~\cite{pramanick2023egovlpv2}. Note that for all evaluations, we removed all Ego4D videos used by the EgoSchema benchmark from our training set to avoid data leakage. Compared to EgoVLP and EgoVLP2, our approach still achieves the best results, outperforming these two baselines by a significant margin of ~16\%, indicating the superiority of our method.

\subsection{Ablation Studies}
\label{sec: ablation}

\noindent\textbf{Ablation of Input Modalities.} Our model utilizes both video features and recursive text inputs (generated in the previous hierarchy) for the segment descriptions and video summaries. Note that we do not use any text inputs for clip captions as they define the base case of our recursive video model. Since we need to sparsely sample video features to fit long-range videos into GPU memory, we hypothesize that using text as an intermediate representation should complement the sparse video features. In Table~\ref{tab: input}, we compare our model with a non-recursive baseline (row 2), which only uses sparse video features and a recursive baseline (row 3), which only uses recursive text features. We observe that combining video and text inputs produces a +1.57\% boost relative to video-only and a +1.64\% boost compared to text-only baselines in CIDEr for segment description generation. Moreover, combining both inputs is more important for long-range video summary generation, where video+text inputs provide +2.42\% and +4.83\% gains compared to video-only and text-only variants. These experiments reveal that the recursive design of \model\ that utilizes both video and text input modalities is crucial for the hierarchical video captioning task.
\begin{table}[t]
\Large
\centering
\resizebox{0.48\textwidth}{!}{%
\begin{tabular}{c|ccc|ccc}
\toprule
\multirow{2}{*}{Input} & \multicolumn{3}{c|}{Segment Description}             & \multicolumn{3}{c}{Video Summary}                \\ \cline{2-7} 
                       & C              & R              & M              & C              & R              & M              \\ \toprule
Video                  & 40.17          & 38.65          & 17.59          & 25.64          & 29.61          & 13.57          \\
Text                   & 40.10          & 38.02          & 17.41          & 23.23          & 29.17          & 13.31          \\
Video + Text           & \textbf{41.74} & \textbf{39.04} & \textbf{18.21} & \textbf{28.06} & \textbf{32.27} & \textbf{14.26} \\ \toprule
\end{tabular}
}
\vspace{-4mm}
\caption{\textbf{Video-Language Input Ablation.} Using both sparse video features and recursive text inputs leads to better performance for both segment description and video summary generation.\vspace{-0.2cm}}
\label{tab: input}
\end{table}

\begin{table}[t]
\LARGE
\centering
\resizebox{0.5\textwidth}{!}{%
\begin{tabular}{c|ccc|ccc}
\toprule
\multicolumn{1}{c|}{\multirow{2}{*}{Training Scheme}} & \multicolumn{3}{c|}{Segment Description}             & \multicolumn{3}{c}{Video Summary}                \\ \cline{2-7} 
\multicolumn{1}{l|}{}                                 & C              & R              & M              & C              & R              & M              \\ \toprule
Init $\rightarrow$ Segment                                & 36.81          & 38.70          & 17.17          & -              & -              & -              \\
Caption $\rightarrow$ Segment                                     & \textbf{41.74} & \textbf{39.04} & \textbf{18.21} & -              & -              & -              \\ \bottomrule
Init $\rightarrow$ Video                               & -              & -              & -              & 8.62           & 26.33          & 11.24          \\
Caption $\rightarrow$ Video                                     & -              & -              & -              & 24.84          & 30.74          & 13.25          \\
Caption $\rightarrow$ Segment $\rightarrow$ Video                     & - & - & - & \textbf{28.06} & \textbf{32.27} & \textbf{14.26} \\ \bottomrule
\end{tabular}
}
\vspace{-0.3cm}
\caption{\textbf{Hierarchical Curriculum Learning.} Using the proposed curriculum learning scheme yields a performance boost of +4.93\% in segment description and +19.44\% in long-range video summary generation compared to training the model from GPT2 pretrained weights (Init).\vspace{-5mm}}
\label{tab: initialization}
\end{table}

\vspace{1mm}
\noindent\textbf{Significance of Hierarchical Curriculum Learning.} Next, we investigate the significance of our hierarchical curriculum learning scheme. Table~\ref{tab: initialization} shows the importance of such a curriculum learning scheme. We observe that if we directly train our model on the segment description from GPT2 pretrained initialization, performance drops by a significant margin of 4.93\% CIDEr. Moreover, the performance drop is even more catastrophic (-19.44\%) for video summary generation without curriculum learning. Finally, we show that it is useful to progressively incorporate higher-level captions, starting from short-term captions, then transitioning to medium-length segment descriptions, and lastly, finishing with long-range video summaries. The variant that progresses from short-term caption to long-range video summary learning directly exhibits a 3.22\% drop in CIDEr performance.

\begin{table}[t]
\centering
    \begin{subtable}{1\linewidth}
    \centering
    \resizebox{1\textwidth}{!}{%
    \begin{tabular}{c|ccc|ccc}
    \toprule
    \multicolumn{1}{c|}{\multirow{2}{*}{LLM}} & \multicolumn{3}{c|}{Segment Description}              & \multicolumn{3}{c}{Video Summary}                                  \\ \cline{2-7} 
    \multicolumn{1}{l|}{}                       & C               & R              & M              & C                    & R                    & M                    \\ \toprule
    GPT2                                        & 96.47           & 46.96          & 23.13          & 40.06                & 33.06                & 14.76                \\
    GPT2-L                                      & 104.30          & 47.68          & 23.15          & 43.18                & 33.86                & 15.00                \\
    FLAN-T5-S                                   & 95.61           & 46.16          & 22.30          & 43.27                & 34.19                & 14.69                \\
    FLAN-T5-L                                   & \textbf{125.67} & \textbf{50.61} & \textbf{26.06} & \textbf{52.08}       & \textbf{36.99}       & \textbf{19.93}       \\ \bottomrule
    \end{tabular}      
    }
    \caption{Training an LLM Teacher.}\vspace{0mm}
    \label{tab: llm oracle}
    \end{subtable}
    \begin{subtable}{1\linewidth}
    \centering
    \resizebox{1\textwidth}{!}{%
    \begin{tabular}{c|ccc|ccc}
    \toprule
    \multirow{2}{*}{
        \begin{tabular}{c}
             Pseudo\\Ann.
        \end{tabular}}                               & \multicolumn{3}{c|}{Segment Description}             & \multicolumn{3}{c}{Video Summary}                \\ \cline{2-7}  & C              & R              & M              & C              & R              & M              \\ \toprule
    \xmark                                                   & 41.74          & 39.04          & 18.21          & 28.06          & 32.27          & 14.26          \\
    \cmark & \textbf{46.88} & \textbf{39.73} & \textbf{18.55} & \textbf{29.34} & \textbf{32.64} & \textbf{14.45} \\ \bottomrule
    \end{tabular}
    }
    \caption{Supervision Using the best LLM Teacher (FLAN-T5-Large).}
    \label{tab: llm supervision}
    \end{subtable}
    \vspace{-0.6cm}
    \caption{\textbf{Importance of LLM Supervision.} \textbf{Top:} Given ground-truth short-term captions concatenated across varying temporal lengths, FLAN-T5-Large generates the highest quality pseudo-annotations for segment description and long-range video summary annotations. Using this LLM Oracle, we produce 100K pseudo-annotations for medium-length segment descriptions and 15K for long-range video summaries. \textbf{Bottom:} Combining LLM-generated annotations with manual annotations during training leads to a performance improvement of 5.14\% CIDEr for segment description and 1.28\% CIDEr for the video summary.\vspace{-5mm}}
    \label{tab: llm}
\end{table}
\vspace{1mm}
\noindent\textbf{Importance of LLM-Based Supervision.} Finally, we study the importance of LLM-based supervision for medium-length segment descriptions and long-range video summaries. In Table~\ref{tab: llm oracle}, we show the performance of different LLM Teachers (e.g., GPT2~\cite{radford2019language}, and FLAN-T5~\cite{chung2022scaling}) that we use to generate the pseudo ground truth data. We observe that FLAN-T5-Large achieves the best performance in all metrics. Hence, we use FLAN-T5-Large as our Teacher to generate pseudo-ground truth data for segment descriptions and long-range video summaries. Specifically, we produce 100K pseudo-annotations for segment descriptions and 15K for video summaries. We combine these pseudo-annotations with the manually annotated data and train our model. Table~\ref{tab: llm supervision} shows that utilizing supervision from LLMs provides a substantial performance boost in both segment description (+5.14\% CIDEr gain) and video summary (+1.28\% CIDEr improvement) generation performance.

\subsection{Qualitative Results on \dataset}
\label{sec:sup qualitative hcap}

\begin{figure*}
    \centering
    \begin{subfigure}{1\textwidth}
        \resizebox{1\textwidth}{!}{%
        \includegraphics[width=1\linewidth]{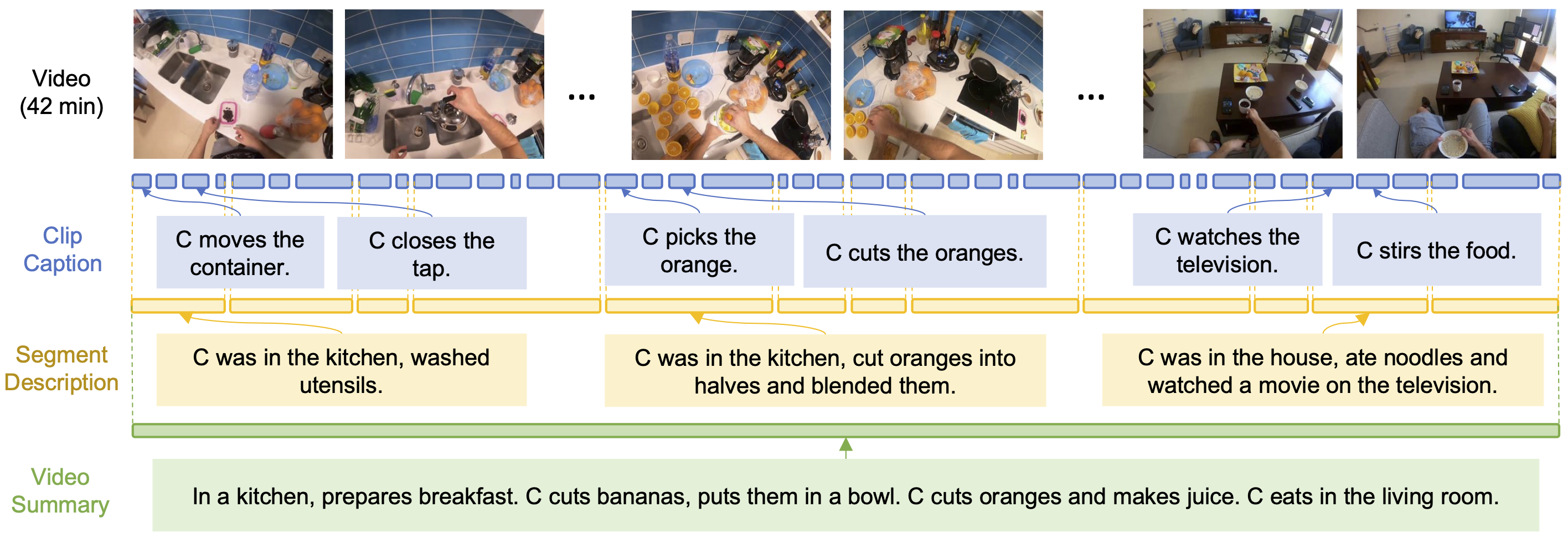}
        }
        \caption{}
    \end{subfigure}
    \begin{subfigure}{1\textwidth}
        \resizebox{1\textwidth}{!}{%
        \includegraphics[width=1\linewidth]{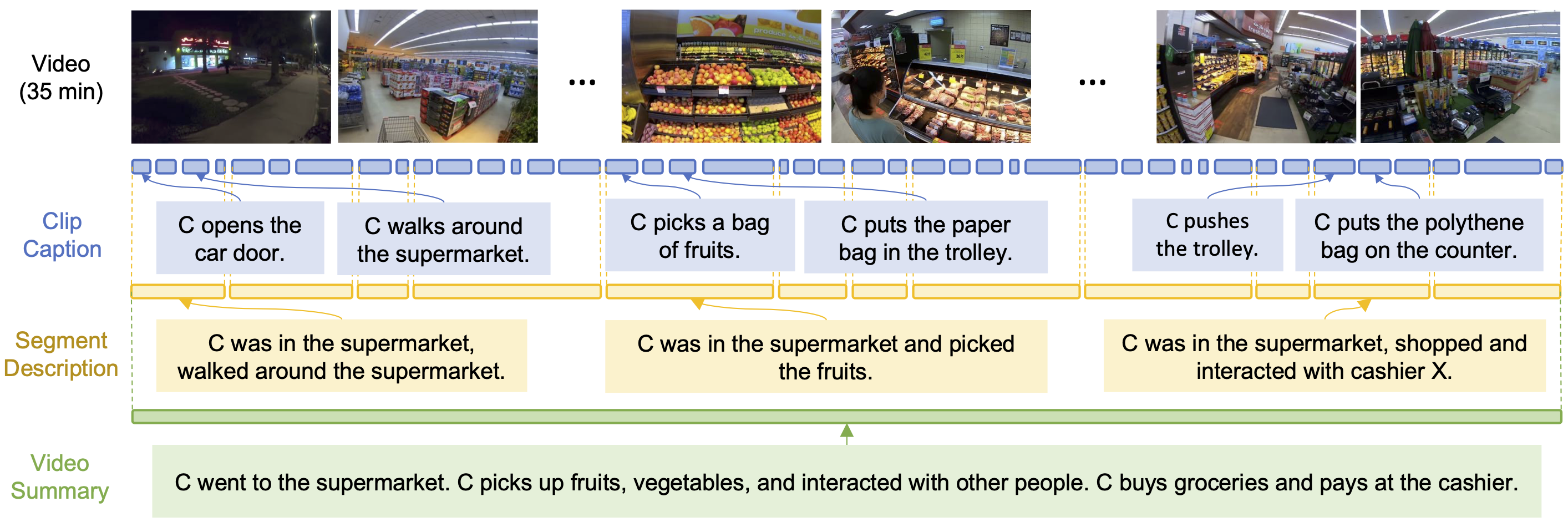}
        }
        \caption{}
    \end{subfigure}
    \begin{subfigure}{1\textwidth}
        \resizebox{1\textwidth}{!}{%
        \includegraphics[width=1\linewidth]{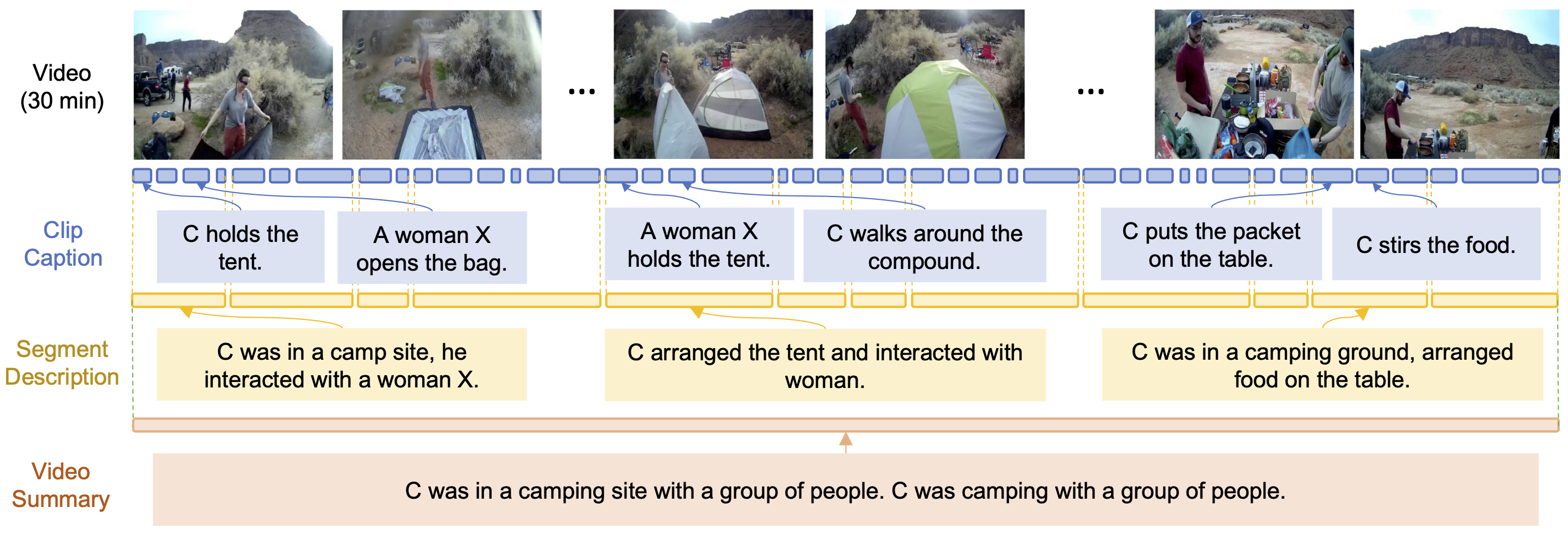}
        }
        \caption{}
    \end{subfigure}
    \caption{\textbf{Qualitative Results on \dataset~.} Generally, clip captions depict atomic actions and objects; segment descriptions focus on intermediate concepts, and video summaries encapsulate the overall content and goals of the videos. While generating clip captions and segment descriptions are often relatively easier tasks, developing a good video summary is often challenging. Our models perform well on video summaries (a) and (b), but the generated video summary (c) could be further improved.}
    \label{fig:example outputs} 
\end{figure*}

In \Cref{fig:example outputs}, we present three instances of hierarchical captions generated by our model. It is evident that clip captions mostly describe atomic actions and objects, such as `C closes the tap' (\Cref{fig:example outputs} (a)) and `C pushes the trolley' (\Cref{fig:example outputs} (b)). In contrast, segment descriptions focus on intermediate concepts within the video spanning longer durations, i.e., `C was in the kitchen, washed utensils' (\Cref{fig:example outputs} (a)), and `C arranged the tent and interacted with a woman' (\Cref{fig:example outputs} (c)).
Moreover, video summaries aim to encapsulate the overarching content and events of the video. For example, `C went to the supermarket. C picked up fruits vegetables, and interacted with other people. C bought groceries and paid at the cashier' (\Cref{fig:example outputs} (b)).

We also notice that while generating clip captions and segment descriptions is relatively more straightforward, generating video summaries is more challenging. For instance, while the generated video summaries of \Cref{fig:example outputs} (a) and \Cref{fig:example outputs} (b) are of good quality, the video summary of \Cref{fig:example outputs} (c) could be further improved. The video summary of \Cref{fig:example outputs} (c) fails to capture some important events of the video and includes repeated words and phrases. These challenges highlight the complexity of summarizing content in long-range videos. We anticipate that future advancements and the use of our released data will contribute to the development of more effective methods and models for this demanding task.
\section{Conclusions and Future Work}
\label{sec:discussion}

We introduce \model\, a recursive video captioning model adept at producing hierarchical captions for videos spanning diverse temporal granularities—from brief clip captions to extensive hour-long summaries. The incorporation of a curriculum learning scheme inspired by human psychology and an LLM-based supervision strategy enhances the model's efficacy in tackling the hierarchical video captioning problem. Beyond its primary focus, our model's hierarchical captions also proves advantageous for long-range video question answering. Additionally, the curated \dataset\ dataset will be released, intended to catalyze ongoing progress in video understanding research. Some promising future directions include real-time caption generation, interactive video understanding, and video-based dialoguing.

\noindent\textbf{Acknowledgements.} We thank Feng Cheng, Yan-Bo Lin, Ce Zhang, Yue Yang, and Soumitri Chattopadhyay for their helpful discussions. Authors from UNC Chapel Hill were supported by the NIH Award R01HD11107402, Sony Faculty Innovation award, Laboratory for Analytic Sciences via NC State University, and ONR Award N00014-23-1-2356.

{
    \small
    \bibliographystyle{ieeenat_fullname}
    \bibliography{main}
}

\clearpage
\maketitlesupplementary

\setcounter{section}{0}
\setcounter{figure}{0}
\setcounter{table}{0}

\renewcommand{\thesection}{S\arabic{section}}
\renewcommand{\thetable}{S\arabic{table}}
\renewcommand{\thefigure}{S\arabic{figure}}


Our supplementary materials contain \Cref{sec:sup implementation}: Additional Implementation Details, \Cref{sec:sup dataset collection}: \dataset~Data Collection Process, \Cref{sec:sup dataset analysis}: \dataset~Dataset Analysis, \Cref{sec:sup results}: Additional Quantitative Results, and \Cref{sec:sup qualitative}: Qualitative Results on EgoSchema.

\section{Additional Implementation Details}
\label{sec:sup implementation}
\begin{figure*}
    \centering
    \resizebox{0.9\textwidth}{!}{%
    \includegraphics[width=1\linewidth]{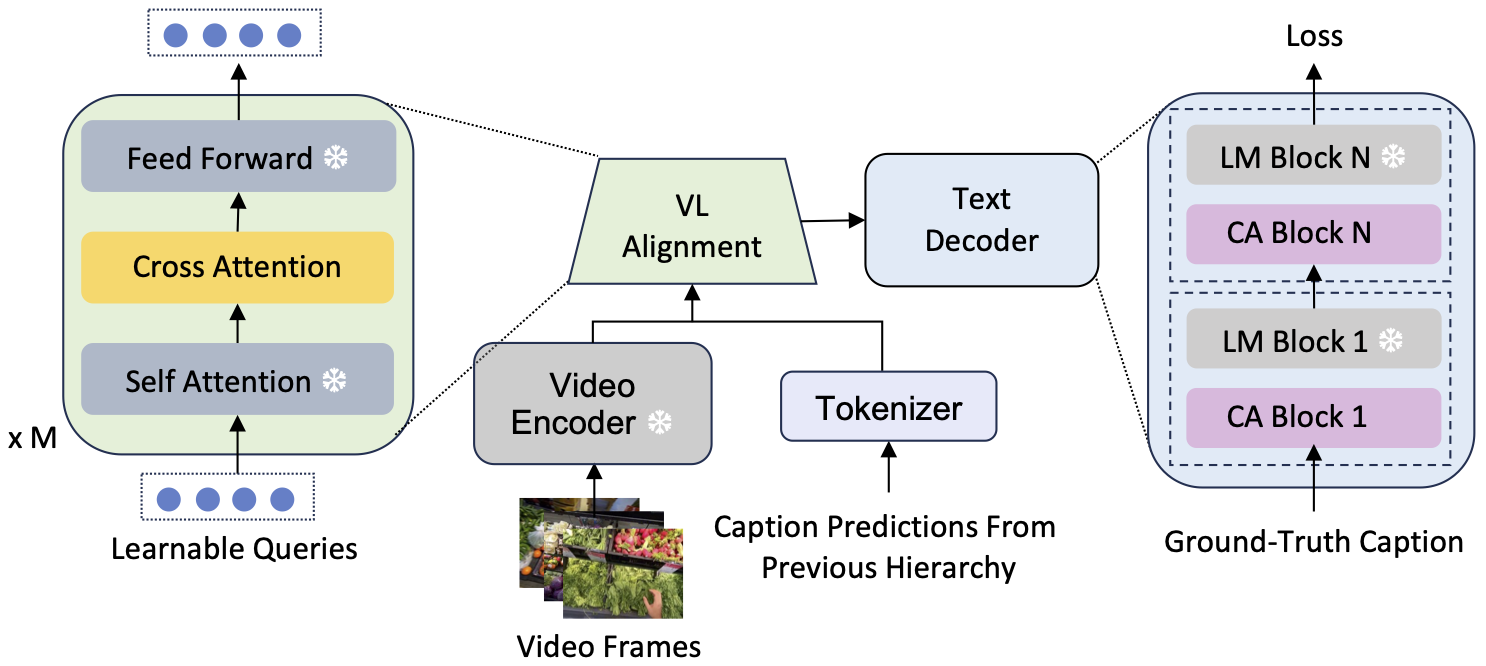}
    }
    \caption{\textbf{Model Architecture.}}
    \label{fig:architecture} 
\end{figure*}

\Cref{fig:architecture} Shows the schematic diagram of the proposed \model~model. 

\noindent\textbf{Video Encoder.} We employ the TimeSformer model~\cite{bertasius2021space} as our video encoder. This model, consisting of 12 transformer layers, is pretrained using a contrastive objective~\cite{zhao2023learning}. The input to the encoder comprises 4 RGB frames of size $224\times224$. To process the video, we divide it into 4-second clips and extract features for each clip using the pretrained video encoder. For clip caption, we utilize the dense spatiotemporal features. This allows our model to capture fine-grained details. However, we only use the CLS features for segment description and video summary, allowing efficient computation.

\noindent\textbf{Video-Language Alignment.} We utilize a pretrained language model DistilBERT~\cite{sanh2019distilbert} as our Video-Language (VL) Alignment module. It is a 6-layer transformer encoder model, where we freeze the self-attention blocks and insert a trainable cross-attention module inside each layer. It takes video features output by the video encoder and captions generated at the previous hierarchy as inputs. Note that there are no text inputs for clip captions. For segment description, we extract clip captions at each 4 seconds of the segment, and for video summary, we extract segment descriptions at each 3 minutes of the video and pass them to the VL alignment module along with corresponding video features. 

\noindent\textbf{Text Decoder.} We leverage a pretrained GPT2~\cite{radford2019language}) as our text decoder. It is a 12-layer transformer model, and we insert a gated cross-attention block inside each transformer layer. We train only the cross-attention modules and freeze the rest of the model. Each cross-attention block contains a cross-attention layer and a feed-forward layer, followed by a tanh gating~\cite{hochreiter1997long}. The tanh-gating is initialized with an initial value of zero so that the model's output is the same as the pre-trained LLM at the beginning. As the training progresses, the model gradually learns to attend to the video-text embedding output by the VL-alignment module.

\noindent\textbf{Training the \model~Model.} We follow a three-stage training pipeline for the \model\ model. First, we train our model 5 epoch using a batch size of 128 using clip caption data, which only uses video features. Afterward, we employ the trained model from the first stage to extract clip captions within the videos at 4-second intervals. Then, during the second stage, we train the model for 10 epochs using a batch size of 32 using segment description samples, which take as input both video features and text features (clip captions). Finally, in the third stage, we extract segment descriptions every three minutes of the video using the trained model of the second stage and further train the model for 10 epochs using a batch size of 32 using video summary data. We use AdamW optimizer with optimizer~\cite{Kingma2014AdamAM} with $(\beta_1 , \beta_2) = (0.9, 0.999)$ and weight decay 0.01. We use a learning rate of $3^{-5}$ and a cosine scheduling strategy.

\noindent\textbf{Training the \umodel~Model.} Training a unified model that shares all parameters across three hierarchies is more challenging. We employ a similar three-stage approach with some additional tricks. In particular, the first-stage training is identical to the \model~model. However, during the second stage, we train the \umodel~model using both clip captions and segment description samples to prevent catastrophic forgetting of clip captions. One particular challenge is that the clip captions and segment description data are quite different. While clip captions use dense spatiotemporal features, segment descriptions utilize CLS features. Moreover, segment descriptions use video and text features as inputs, while clip captions only use video features. To overcome this challenge, we employ an alternate batching pipeline, where we sample a batch of clip captions and segment descriptions alternatively during the training. Since we have a lot more clip caption data ($\sim4M$) compared to segment descriptions ($100K$ including manually annotated and LLM-generated pseudo annotations), we randomly sample $100K$ clip captions and only used those during the second stage of training. Finally, we train the model during the third stage using samples from all three hierarchies using a similar alternate batching approach. Since we have only $\sim20K$ (including manually annotated and LLM-generated pseudo annotations) samples for video summaries, we randomly sample $20K$ clip captions and 20K segment descriptions and used those along with video summaries during the third stage of training. This strategy prevents catastrophic forgetting of the model. It allows the training of the \umodel~model, which shares all parameters across hierarchies. For \umodel, We use the same learning rate, batch size, training epoch, optimizer, and scheduler for the \model~(See the previous paragraph).

\noindent\textbf{Inference.} During inference, we uniformly sample 4 frames from the corresponding clip and extract spatiotemporal features using the video encoder to use as inputs to generate clip captions. For segment description, we extract CLS features and clip captions every 4 seconds of the segment and use them as inputs to generate segment descriptions. Lastly, we extract segment descriptions at each 3 minutes of the video and use them along with pre-extracted CLS features to generate video summaries. Note that clip boundaries are not given during the inference of segment descriptions, and segment boundaries are not given during the inference of video summaries.

We will release our code, data, and pretrained models.

\section{\dataset\ Data Collection Process}
\label{sec:sup dataset collection}

The Ego4D-HCap dataset was collected over the span of 2 months, from April 2023 to May 2023 and from September 2023 to October 2023. We recruited 91 specialized annotators through CloudResearch\footnote{https://www.cloudresearch.com}, a participant-sourcing company. All annotators are based in the United States and are compensated at a rate of 9 dollars per hour, which is above the national minimum wage.

\begin{figure*}
    \centering
    \resizebox{1\textwidth}{!}{%
    \includegraphics[width=1\linewidth]{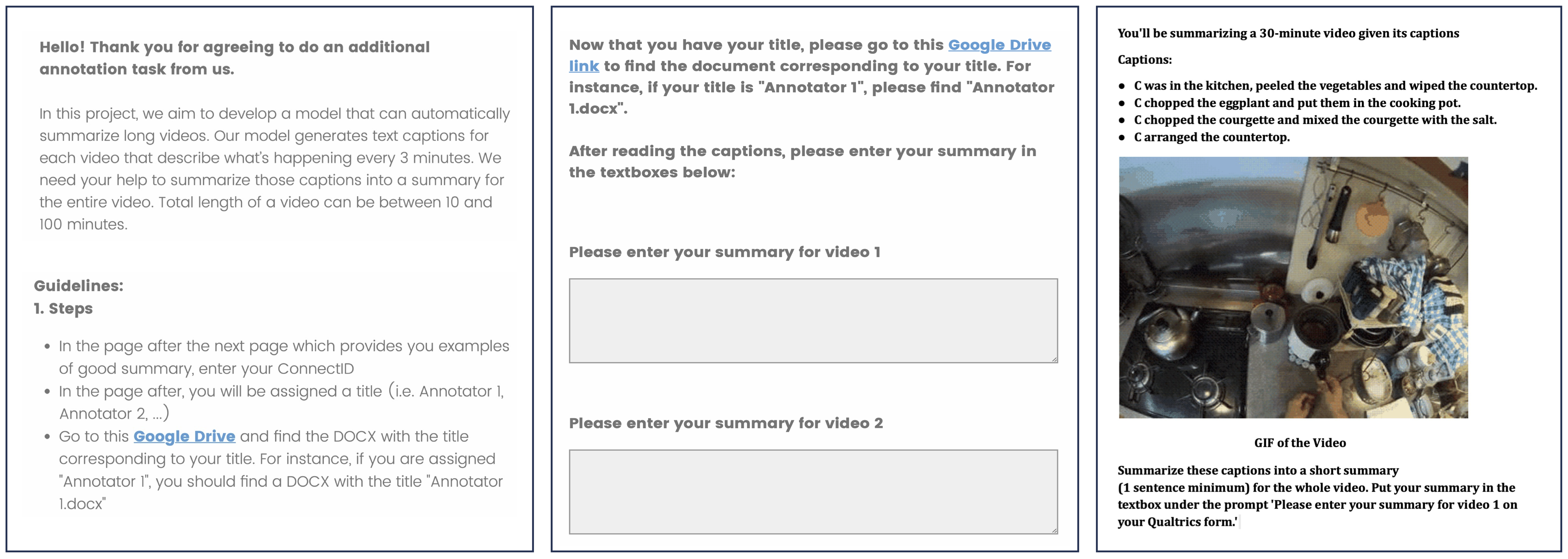}
    }
    \caption{\textbf{Data Collection Interface.}}
    \label{fig:data collection} 
\end{figure*}

We utilized Qualtrics and Google Drive to build our data collection interface. Our interface began with an introduction to our project, guidelines for summarizing the videos, and examples of good summaries. It then asked the annotators for their ConnectID and provided them a link to the documents of videos assigned to them. Each document would contain 10-25 videos for the annotators to summarize, along with a prompt and a GIF summarizing the events of each video. The last interfaces contain text boxes for the annotators to put the text summaries for each video and the annotator's experience with the data collection interface. We used the latter to improve upon the interface so that the quality of the annotated summaries ultimately became better. \Cref{fig:data collection} shows our data collection interface.

\subsection{Guidelines for Annotators}
\label{sec:sup guidelines}

\textbf{Overview.} In this project, we aim to develop a model that can automatically summarize long videos. Our model generates text captions for each video describing what happens every 3 minutes. We need your help to summarize those captions into a summary for the entire video. The total length of a video can be between 10 and 100 minutes. 

\noindent\textbf{Captions.}
\begin{enumerate}
    \item You are given a list of captions for each video. 
    
    \item Each caption describes what is happening every 3 minutes.

    \item C refers to a person in the provided captions.
    
    \item The captions are generated using a machine learning model, so sometimes, they can be out of order or inaccurate. In that case, you can exclude the events or details that do not make sense in the summary or refer to the GIF provided under the captions.
    
    \item The captions may also use different terms to refer to the same thing. If only technical terms are used, then use them in your summary. Otherwise, we prefer you to use generic terms. 
\end{enumerate}

\noindent\textbf{GIFs.}
\begin{enumerate}
    \item Since the videos are very long, we do not provide the full video. Instead, you are also given a GIF for each video. 
    \item GIFs created by sparsely sampled frames from the video, which is intended to help you better understand the overall contents of the video along with the captions.
\end{enumerate}

\label{sec:expectations}
\noindent\textbf{Summaries.}
\begin{enumerate}
    \item The summary should be one paragraph long. Try to maintain a compression factor of 5, i.e., for every five captions, you should summarize it in 1 sentence. However, each summary should be at least one sentence.

    \item The summary should cover the setting, characters, and events that take place in the order of the video.

    \item Avoid using X, Y or other letters to refer to characters other than C. Instead, use woman and man. Refer to examples of good summaries on the next page.

    \item The summary should not have an interpretation of the characters’ personalities or qualities.

    \item The summary should be logically coherent, unambiguous, and understandable.

    \item The summary should be grammatically correct.

    \item Repetition of actions should have an underlying purpose/pattern. 
\end{enumerate}

\subsection{Quality Control}
To control the quality of the annotations, we pre-selected annotators before moving them forward with the official annotation task and manually reviewed the annotations. Before the official annotation task, we paid 171 annotators to complete a preliminary annotation task and selected from this pool annotators who provided desirable annotation quality. We minimized the chances of getting low-quality annotations by pre-selecting high-quality annotators and familiarizing them with an interface similar to the actual annotation task.

Another quality control method we utilized was to review the annotations ourselves manually. For each annotator, we randomly sampled half of the annotations they provided. We assessed their quality based on whether they followed the expectations outlined in \Cref{sec:sup guidelines}. If less than half of the sampled annotations are of low quality, we would provide annotator feedback and ask them to redo their annotations. If the annotations were of better quality, we would replace them with the initial annotation. Otherwise, we would discard both versions and assign them to other annotators.

\subsection{De-identification Process}

Due to the nature of the dataset and our task, our dataset has already been de-identified. Since all of our videos are sourced from Ego4D, they have undergone sensitive object detection, false positive removal, fast negative correction, and image blurring ~\cite{grauman2022ego4d}. They were not modified during the dataset collection process, so the videos remain de-identified. Our annotators are also anonymized, as we recruited, managed, and corresponded with annotators on CloudResearch. Aside from their ConnectID, which we used to revise annotations, we did not collect any of the annotators' personal information.

\section{\dataset\ Dataset Analysis}
\label{sec:sup dataset analysis}

\subsection{Example Video Summaries.} 
\begin{figure*}[b]
    \centering
    \resizebox{1\textwidth}{!}{%
    \includegraphics[width=1\linewidth, height=0.75\linewidth]{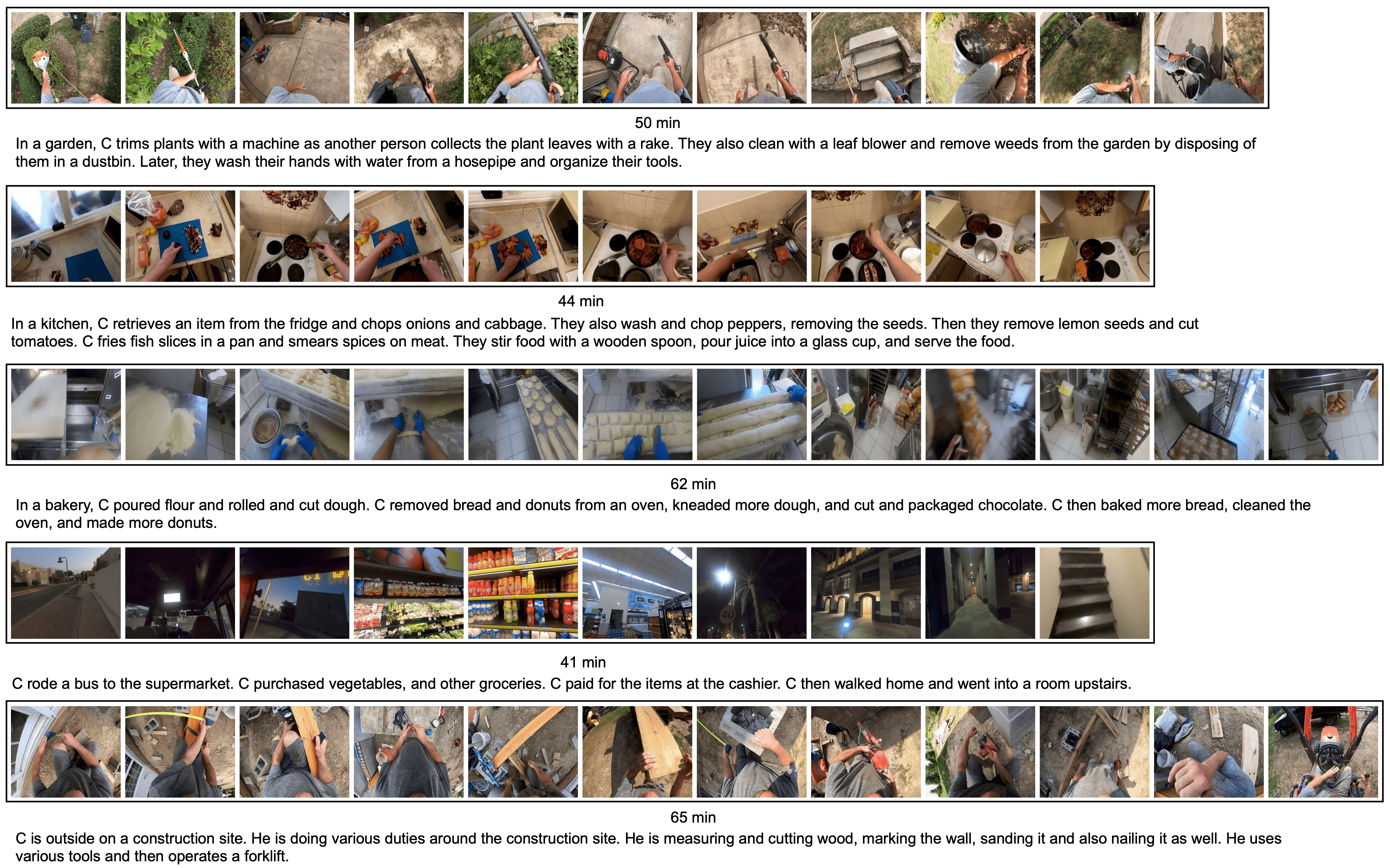}
    }
    \caption{\textbf{Examples of annotated video summaries of the \dataset\ dataset.} Due to space limitation and conciseness, we show one frame for each 5 minutes of the video..}
    \label{fig:example videos} 
\end{figure*}

\Cref{fig:example videos} Shows examples of annotated video summaries of the \dataset\ dataset. We observe that video summaries are of various lengths and capture diverse scenarios, places, and activities. Typically, each video is annotated with multiple summaries. However, the figure shows only one summary per video for clarity and conciseness. 

\section{Additional Quantitative Results}
\label{sec:sup results}

\begin{table*}
\centering
\resizebox{0.8\textwidth}{!}{%
\begin{tabular}{c|c|ccc|ccc|ccc}
\toprule
\multirow{2}{*}{\begin{tabular}[c]{@{}c@{}}LM \\ Alignment\end{tabular}} & \multirow{2}{*}{\begin{tabular}[c]{@{}c@{}}Trainable\\ CA\end{tabular}} & \multicolumn{3}{c|}{Clip Caption}                & \multicolumn{3}{c|}{Segment Description}             & \multicolumn{3}{c}{Video Summary}                \\ \cline{3-11}    &                                                                         & C              & R              & M              & C              & R              & M              & C              & R              & M              \\ \toprule
\xmark                                                                        & \cmark                                                                       & 92.56          & 47.64          & 28.03          & 39.41          & 38.62          & 17.71          & 23.04          & 28.33          & 13.72          \\
\cmark                                                                        & \xmark                                                                       & 73.88          & 43.17          & 21.67          & 32.16          & 31.67          & 13.33          & 12.16          & 21.06          & 8.22           \\
\cmark                                                                        & \cmark                                                                       & \textbf{98.35} & \textbf{48.77} & \textbf{28.28} & \textbf{41.74} & \textbf{39.04} & \textbf{18.21} & \textbf{28.06} & \textbf{32.27} & \textbf{14.26} \\ \bottomrule
\end{tabular}
}
\caption{\textbf{Architecture Ablation.} An LM-based~\cite{sanh2019distilbert} Video Language Alignment module provides significant performance gains compared to the transformer-based resampler used in prior works~\cite{zhao2023learning, alayrac2022flamingo}. Adding trainable cross-attention layers inside the text decoder performs much better than freezing the decoder.}
\label{tab: architecture}
\end{table*}

\noindent\textbf{Backbone Design.} In this section, we ablate various aspects of our Video-Language Backbone design. First, we validate the effectiveness of a Language Model-based (LM)~\cite{sanh2019distilbert} Video-Language Alignment module rather than a standard Transformer resampler used in prior works~\cite{zhao2023learning, alayrac2022flamingo}. Table~\ref{tab: architecture} shows that an LM-based Alignment module performs significantly better than the standard transformer-based resampler in all three hierarchies. Second, we inject trainable cross-attention layers~\cite{zhao2023learning, alayrac2022flamingo} in the text decoder to incorporate video features. In contrast, several prior works~\cite{li2023blip, mokady2021clipcap} inject video features only in the input layer while freezing the whole text decoder. \Cref{tab: architecture} shows that using trainable cross-attention layers in the textual decoder performs significantly better than using video features in the input layer alone across all three hierarchical levels.

\section{Qualitative Results on EgoSchema}
\label{sec:sup qualitative}

\begin{figure*}
    \centering
    \vspace{-5mm}
    \begin{subfigure}{1\textwidth}
        \resizebox{1\textwidth}{!}{%
        \includegraphics[width=1\linewidth]{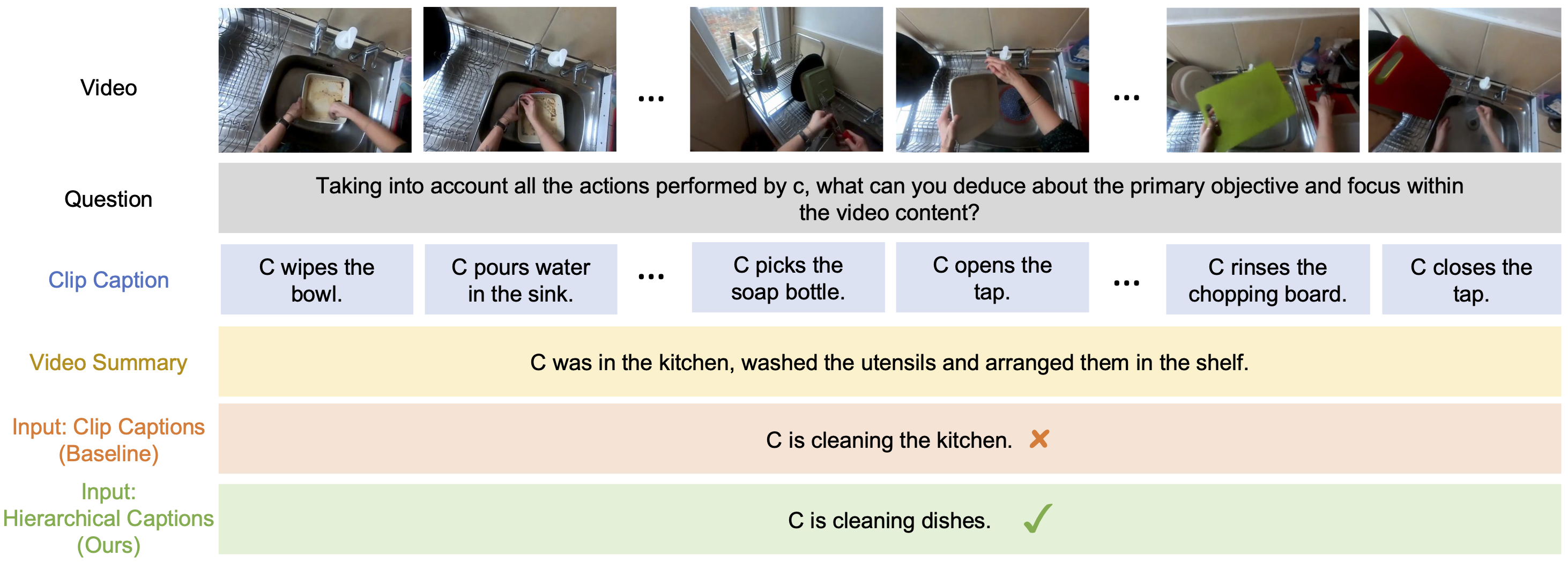}
        }
        \caption{}
    \end{subfigure}
    \begin{subfigure}{1\textwidth}
        \resizebox{1\textwidth}{!}{%
        \includegraphics[width=1\linewidth]{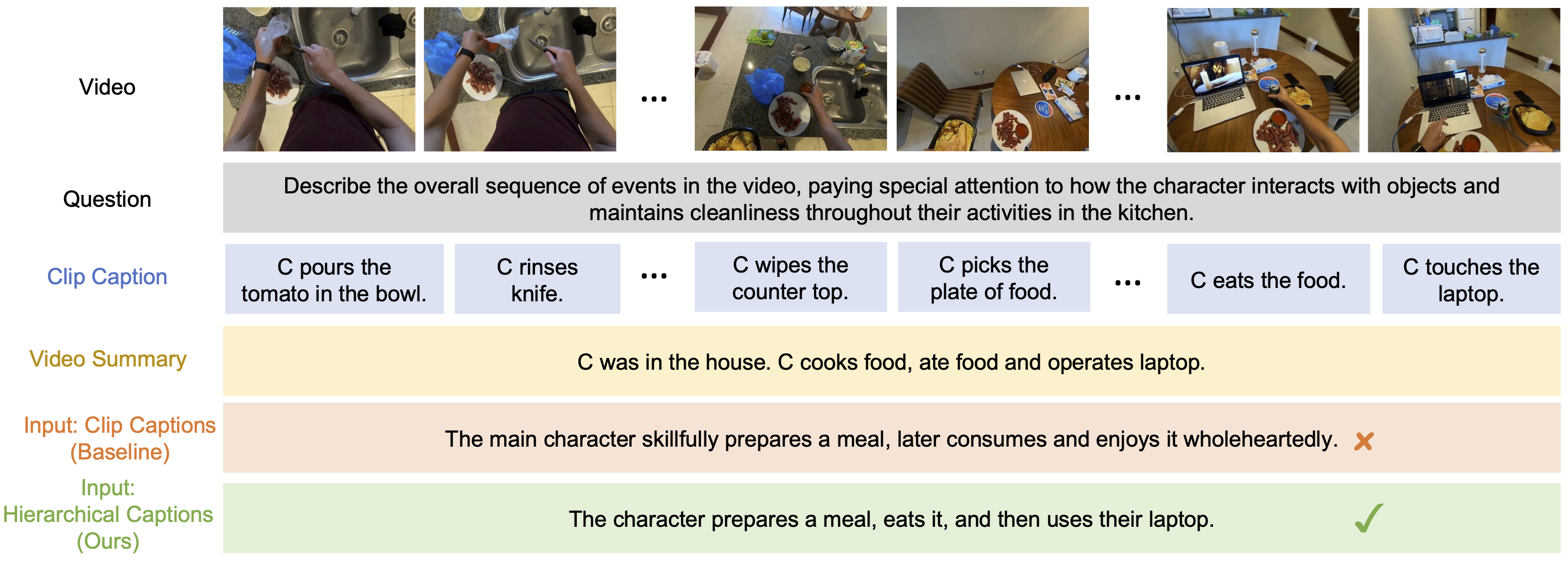}
        }
        \caption{}
    \end{subfigure}
    \begin{subfigure}{1\textwidth}
        \resizebox{1\textwidth}{!}{%
        \includegraphics[width=1\linewidth]{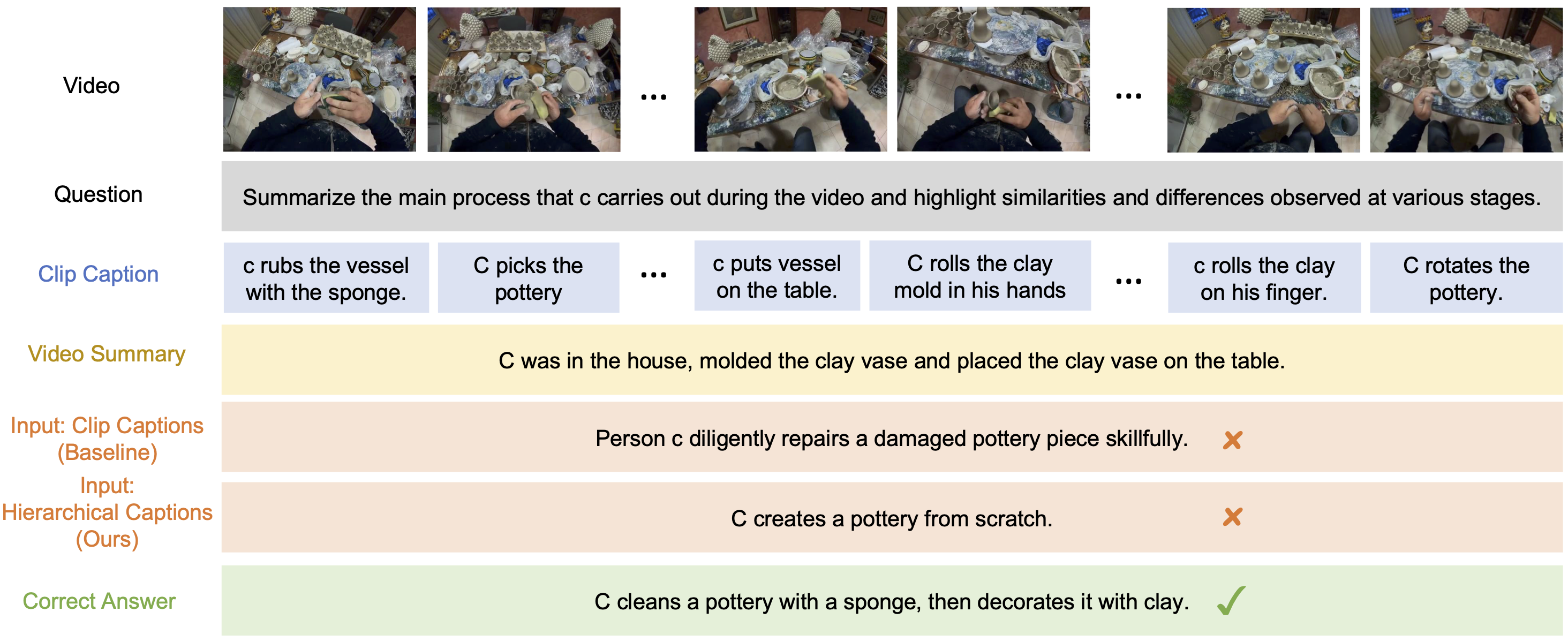}
        }
        \caption{}
    \end{subfigure}
    \caption{\textbf{Qualitative Results on EgoSchema.} The baseline method that uses only short-range clip captions as input fails in examples (a) and (b), where our approach succeeds by utilizing hierarchical captions (i.e., clip captions and video summaries). Both models fail in Example (c).}
    \label{fig:example es} 
\end{figure*}

\Cref{fig:example es} illustrates the qualitative outcomes of our long-range video question answering experiment on the EgoSchema~\cite{mangalam2023egoschema} dataset. The approach, detailed in \Cref{sec:results es}, involves the generation of hierarchical captions utilizing the \model\ model for videos. Subsequently, these captions are presented to ChatGPT along with questions and answer choices as prompts, enabling the model to select the correct answer. In \Cref{fig:example es} (a) and \Cref{fig:example es} (b), it is evident that ChatGPT tends to choose incorrect answers when provided solely with clip captions. However, the model consistently makes correct choices in both scenarios when supplemented with video summaries. This highlights the efficacy of our generated hierarchical captions in enhancing the performance of long-range video question answering tasks. Nevertheless, in certain instances, as depicted in \Cref{fig:example es} (c), our approach encounters challenges and fails to identify the correct answer.

\end{document}